\documentclass[lettersize,journal]{IEEEtran}
\usepackage{amsmath,amsfonts}
\usepackage{algorithmic}
\usepackage{algorithm}
\usepackage{array}
\usepackage[caption=false,font=normalsize,labelfont=sf,textfont=sf]{subfig}
\usepackage{textcomp}
\usepackage{stfloats}
\usepackage{url}
\usepackage{enumerate}
\usepackage{verbatim}
\usepackage{graphicx}
\usepackage{cite}
\usepackage{amsmath}
\usepackage{amsmath, bm}
\usepackage{bbding}
\usepackage{pifont}
\usepackage{booktabs}
\usepackage{amsfonts}
\usepackage{hyperref}
\usepackage{multirow}
\usepackage{dsfont}
\usepackage{color}
\usepackage[table]{xcolor}
\usepackage{bbm}
\usepackage{textcomp}
\begin{document}
%titile
\title{Not just Learning from Others but Relying on Yourself: A new perspective on Few-Shot Segmentation in Remote Sensing}
%author

\author{
 Hanbo~Bi,~%\IEEEmembership{Member,~IEEE,}
 Yingchao~Feng,~\IEEEmembership{Member,~IEEE,}
 Zhiyuan~Yan,~\IEEEmembership{Member,~IEEE,}\\
 Yongqiang~Mao,~\IEEEmembership{Graduate Student Member,~IEEE,}
 Wenhui~Diao,~\IEEEmembership{Member,~IEEE,}\\
 Hongqi~Wang,~\IEEEmembership{Member,~IEEE,}
 and~Xian~Sun,~\IEEEmembership{Senior Member,~IEEE}
        % <-this % stops a space
\thanks{H. Bi and Y. Feng contribute equally to this work.}
\thanks{This work for Few-Shot segmentation in Remote Sensing was supported by the National Key R{\&}D Program of China (2022ZD0118402), the National Natural Science Foundation of China (NSFC) under Grant 62171436, and under Grant 62301538. (\emph{Corresponding author: Wenhui Diao.})} 

\thanks{H. Bi, Y. Mao, X. Sun are with the Aerospace Information Research Institute, Chinese Academy of Sciences, Beijing 100190, China, also with the School of Electronic, Electrical and Communication Engineering, University of Chinese Academy of Sciences, Beijing 100190, China, also with the University of Chinese Academy of Sciences, Beijing 100190, China, and also with the Key Laboratory of Network Information System Technology (NIST), Aerospace Information Research Institute, Chinese Academy of Sciences, Beijing 100190, China (e-mail: bihanbo21@mails.ucas.edu.cn; maoyongqiang19@mails.ucas.ac.cn; sunxian@aircas.ac.cn).}
 
\thanks{Y. Feng, Z. Yan, W. Diao and H. Wang are with the Aerospace Information Research Institute, Chinese Academy of Sciences, Beijing 100190, China, and also with the Key Laboratory of Network Information System Technology (NIST), Aerospace Information Research Institute, Chinese Academy of Sciences, Beijing 100094, China (e-mail: fengyingchao17@mails.ucas.edu.cn; yanzy@aircas.ac.cn; diaowh@aircas.ac.cn; wiecas@sina.com).}}

% The paper headers
\markboth{Journal of \LaTeX\ Class Files,~Vol.~14, No.~8, August~2021}%
{Shell \MakeLowercase{\textit{et al.}}: A Sample Article Using IEEEtran.cls for IEEE Journals}

% \IEEEpubid{0000--0000/00\$00.00~\copyright~2021 IEEE}
% Remember, if you use this you must call \IEEEpubidadjcol in the second
% column for its text to clear the IEEEpubid mark.

\maketitle

\begin{abstract}\label{abstract}
Few-shot segmentation (FSS) is proposed to segment unknown class targets with just a few annotated samples. Most current FSS methods follow the paradigm of mining the semantics from the support images to guide the query image segmentation. However, such a pattern of `learning from others' struggles to handle the extreme intra-class variation, preventing FSS from being directly generalized to remote sensing scenes. To bridge the gap of intra-class variance, we develop a Dual-Mining network named DMNet for cross-image mining and self-mining, meaning that it no longer focuses solely on support images but pays more attention to the query image itself. Specifically, we propose a Class-public Region Mining (CPRM) module to effectively suppress irrelevant feature pollution by capturing the common semantics between the support-query image pair. The Class-specific Region Mining (CSRM) module is then proposed to continuously mine the class-specific semantics of the query image itself in a `filtering' and `purifying' manner. In addition, to prevent the co-existence of multiple classes in remote sensing scenes from exacerbating the collapse of FSS generalization, we also propose a new Known-class Meta Suppressor (KMS) module to suppress the activation of known-class objects in the sample. Extensive experiments on the iSAID and LoveDA remote sensing datasets have demonstrated that our method sets the state-of-the-art with a minimum number of model parameters. Significantly, our model with the backbone of Resnet-50 achieves the mIoU of 49.58{\%} and 51.34{\%} on iSAID under 1-shot and 5-shot settings, outperforming the state-of-the-art method by 1.8{\%} and 1.12{\%}, respectively. The code is publicly available at \href{https://github.com/HanboBizl/DMNet/}{https://github.com/HanboBizl/DMNet/}.
\end{abstract}


\section{Introduction}\label{Indroduction}
\IEEEPARstart{S}{emantic} segmentation is an essential fundamental task in the intelligent interpretation for remote sensing, aiming at assigning each pixel to one of the pre-defined classes of geographical targets \cite{dey2010review,yuan2021review,mao2022beyond,feng2021double,mao2023light}. It is currently applied in a wide variety of fields, including urban management \cite{mao2023elevation,weng2018urban}, land-use and land-cover mapping \cite{liu2017classifying,gui2022infrared}, environmental monitoring \cite{yuan2020deep}, change detection \cite{wang2022change,wangchange}, and road extraction \cite{zhang2018road,wang2022ddu}. 

Recently, semantic segmentation tasks have made encouraging progress \cite{chen2014semantic,long2015fully,ronneberger2015u,feng2021continual} due to the growing wave of deep learning \cite{goodfellow2016deep,lecun2015deep}, particularly with the emergence of convolutional neural networks (CNNs). Yet, the success of classical CNNs relies heavily on large-scale labeled data, and when labeled data is insufficient or novel (unknown) classes arrive, their performance dramatically degrades \cite{gavrilov2018preventing}. Obtaining such large-scale labeled data is time-consuming and labor-intensive, especially for remote sensing images that are difficult to interpret \cite{wang2020adaptive,wang2022hyperspectral,li2022graph}. Even though various methods have been suggested to alleviate the annotation problem, such as semi-supervised and weakly supervised learning, a need for extensive or weak annotations remains unchanged \cite{zhou2018brief,zhou2021semi}. Furthermore, directly parsing unknown classes tends to be ineffective due to limitations in training paradigms and generalization capabilities, while fine-tuning for unknown classes is time-consuming and laborious, preventing it from satisfying the need for rapid deployment \cite{wang2021dmml,cheng2022holistic}. 

\begin{figure}[t]
	\setlength{\abovecaptionskip}{1pt}
	\centering
	\includegraphics[width=1.0\linewidth]{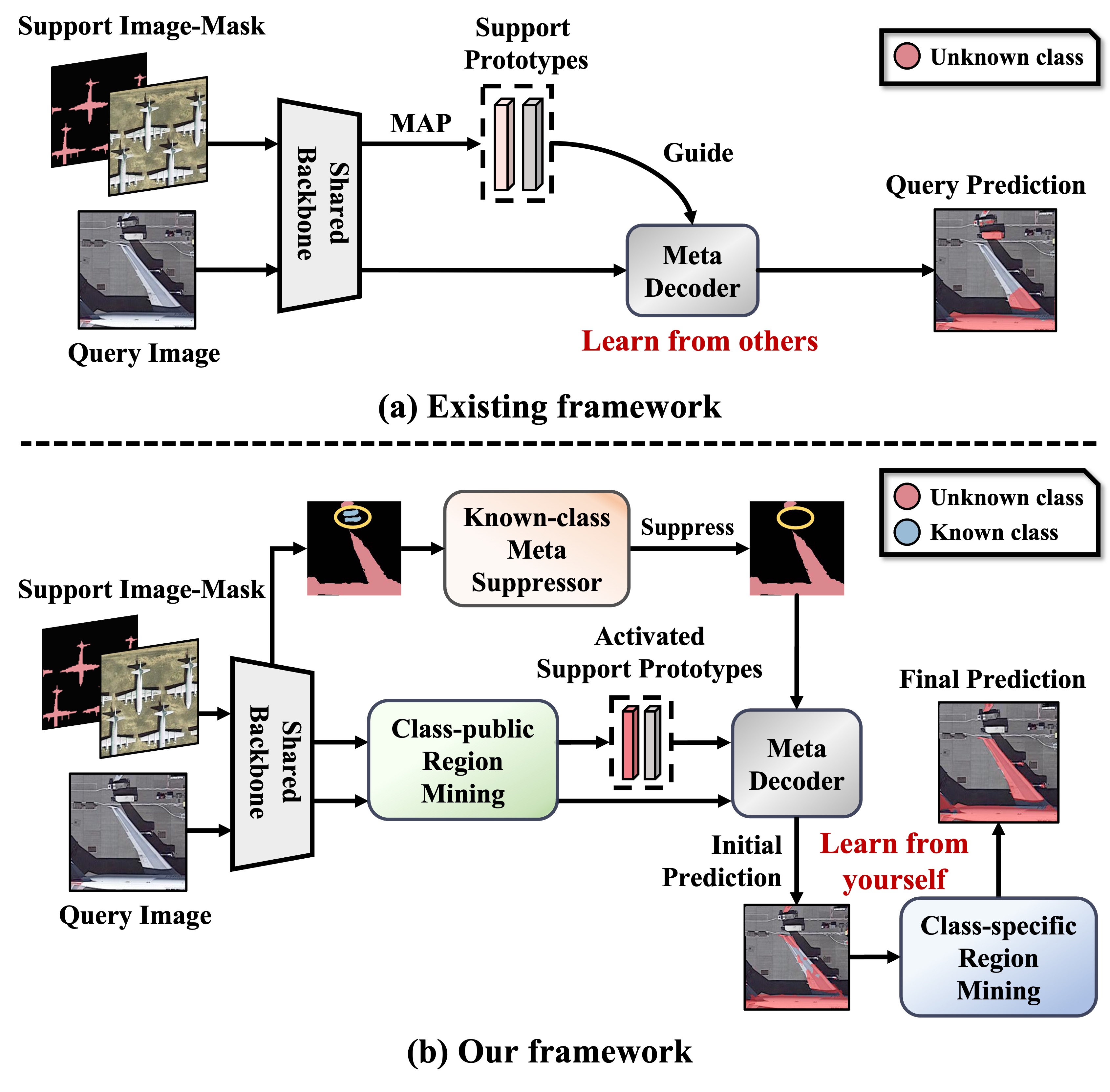}
	\caption{\textbf{Comparison between existing framework and ours.} Existing methods are devoted to mining the support image but ignore the gap of intra-class variance. We mine not only the semantics from the support image but also the semantics from the query image itself for segmentation. In addition, a novel suppressor is introduced to suppress the over-fitting of known classes (e.g., `large vehicle').}
	\label{fig:0}
\end{figure}

\begin{figure}[t]
	\setlength{\abovecaptionskip}{1pt}
	\centering
	\includegraphics[width=0.8\linewidth]{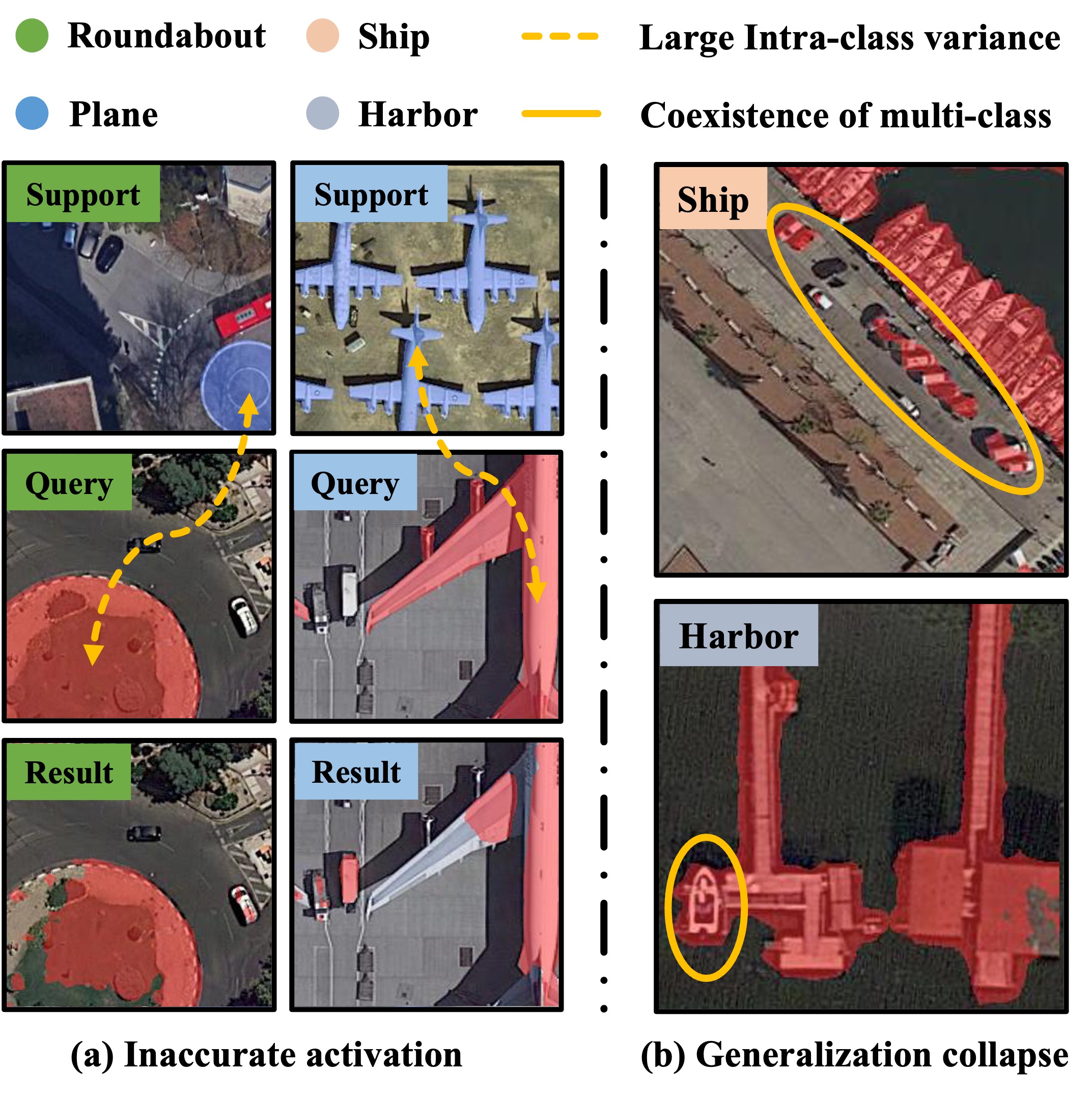}
	\caption{\textbf{The challenges faced by FSS in remote sensing scenarios.} (a) Large intra-class variations in the remote sensing scene can lead to inaccurate activation, where the \textbf{dotted line} represents the difference between the support-query image pair. (b) The co-existence of multi-class in the remote sensing scene can exacerbate the collapse of the generalization, where the \textbf{solid line} represents the false activation of irrelevant classes. }
	\label{fig:1}
\end{figure}

Inspired by the ease with which humans can quickly identify new concepts or patterns from just a few examples, some researchers have suggested Few-shot learning (FSL) to address the above challenges \cite{finn2017model}. FSL methods typically follow the training paradigm of meta-learning, i.e., learning transferable meta-knowledge from known classes (training classes) and generalizing to unknown classes with just a handful of annotated samples. Few-shot segmentation (FSS) is an application of FSL to the semantic segmentation task, which aims to segment unknown class targets in query images utilizing just a handful of annotated samples named support images. \cite{shaban2017one,zhang2019canet,tian2020prior,lang2022beyond,zhang2020sg}. Fig.\ref{fig:0}(a) illustrates the current mainstream framework for FSS: they extract the feature representations (i.e., support prototypes) of the target class from the support images through the masked average pooling (MAP) operation \cite{zhang2020sg} and utilize the support prototypes in a certain way via a meta-decoder to activate target regions in the query image to complete the segmentation.
Although remarkable progress has been made in natural scenes, extreme intra-class variance and multi-class co-existence prevent FSS from being directly generalized to complex remote sensing scenarios, as shown in Fig.\ref{fig:1}. \textbf{(a) Large intra-class variance leads to inaccurate activation}. Extreme intra-class variation in remote sensing scenes makes a huge difference between support and query images (e.g., `roundabouts' and `planes'), where the paradigm of utilizing the support images to guide segmentation fails to provide sufficiently effective guidance and hence is sub-optimal. As shown in Fig.\ref{fig:1}(a), utilizing the support image `bombers' to guide the query image `airliner' only activates the similar `fuselage' parts, while the dissimilar `wing' parts fail. Meanwhile, the direct masked average pooling (MAP) operation will inevitably introduce category semantics specific to the support image itself, which is detrimental to the guidance, i.e., the derived `wing' semantics from the `bomber' mismatch the `airliner'. \textbf{(b) Multi-class co-existence exacerbates the collapse of generalization}. Meta-training with a huge amount of known class data will inevitably introduce the knowledge of known classes and produce over-fitting, hindering generalization to unknown classes. Such generalization collapse is more likely to occur in remote sensing scenarios where multiple classes are prone to co-exist in top-down camera angles (i.e., the existence of irrelevant categories), and such irrelevant categories can easily be activated incorrectly due to the over-fitting of known classes. For example, in Fig.\ref{fig:1}(b), the known classes `small vehicles' and `ships' are incorrectly activated in segmenting the unknown classes `ship' and `harbor', respectively.
To address the above two problems, we suggest a unique Dual-Mining network named \textbf{DMNet} for cross-image mining and self-mining, which consists of three efficient modules: Class-public Region Mining (CPRM), Class-specific Region Mining (CSRM), and Known-class Meta Suppressor (KMS). Fig.\ref{fig:0}(b) briefly describes the roles of each module in DMNet. 

Specifically, we argue that each image is unique and cannot be achieved merely by `imitating' and `borrowing' from other images, i.e., simply relying on the category semantics from the support image to guide the segmentation is not sufficient; one also needs to mine the query image itself. Following this idea, the CPRM and CSRM modules are proposed to capture the public semantics between the image pair and specific semantics from the query image itself, respectively, thus bridging the gap of intra-class variation. \textbf{(a) Take its essence and discard its dross}. First, as with the previous methods, we need to learn relevant knowledge from others (support images) to benefit ourselves (query image). To suppress useless semantic pollution introduced from support images through the MAP operation, the CPRM module focuses on capturing the bidirectional semantic associations between the support-query image pair and activating semantics with strong associations. Thus, common semantics between the image pair are activated while inconsistent parts are weakened. Compared with the support semantics that retain more specific information about the support image, these activated common semantics are closer to the query features in the semantic space, which are named class-public semantics in this paper.
\textbf{(b) Pay more attention to yourself}. It is sub-optimal to utilize the semantics of the support images to guide the query image for segmentation when faced with complex remote sensing scenarios, as it fails to truly bridge the gap of intra-class variance. Thus, the CSRM module starts with itself, mining the category semantics from the query image itself to guide its segmentation. We argue that there are latent category semantics in the prediction result derived from segmenting the query image by support images that can be mined and exploited. Based on this, we propose to filter and purify the prediction result to derive relevant category semantics and then exploit them in a continuously guided manner to activate other target semantics in the query image. Such category semantics from the query image itself match the query image better in the semantic space compared to the semantics in the support image. In this paper, we name these category semantics from the query image as class-specific semantics.

In addition, to address the generalization collapse for unknown classes due to the meta-training paradigm in remote sensing scenarios, we offer a unique meta-training strategy and the corresponding Known-class Meta Suppressor (KMS). Instead of focusing on how to minimize the bias toward knowledge of known classes, we choose to exploit the knowledge to suppress the activation of known classes in the query image. In particular, unlike the traditional meta-training paradigm, the proposed meta-training paradigm sets up an additional branch to continuously learn the semantics of the known classes and store them in a meta-memory in the form of prototypes. During the testing phase, the KMS module utilizes the representative prototypes of the known classes from the meta-memory and the support prototypes of the target class to guide the segmentation jointly. Concretely, we regard the activated high-confidence regions of the known classes as the background so as to achieve the suppression of known classes in the query image. For example, the meta-activation map in Fig.\ref{fig:0}(b) suppresses the activation of the known class `large vehicle'.

Based on the three modules mentioned above, the proposed DMNet can effectively handle complex remote sensing scenarios. Extensive experiments on two remote sensing datasets, iSAID and LoveDA, have demonstrated that our DMNet achieves new state-of-the-art performance under all settings.
The main contributions of our work can be summarized as follows:

\begin{enumerate}
    \item A novel FSS framework is suggested that no longer focuses solely on the knowledge from the support images but pays more attention to the query image itself, providing new insights for future work.
    \item The proposed CPRM module explicitly captures the common semantics and weakens irrelevant semantics between the image pair to suppress useless feature pollution from the support images. 
    \item The proposed CSRM module focuses on mining the specific semantics from the query image itself to guide its segmentation, effectively bridging the gap of large intra-class variance.
    \item To alleviate the collapse of generalization in complex remote sensing scenes, we construct a unique KMS module that continuously learns knowledge of known classes during the training phase to suppress the activation of known-class objects in the sample.
\end{enumerate}

\section{Related work}\label{Related work}
\subsection{Semantic Segmentation}
Semantic segmentation aiming at predicting pixel-wise labels in images is a fundamental task in computer vision \cite{wang2021transformer}. The appearance of CNNs has made the semantic segmentation task a hot topic. Full convolutional network (FCN) \cite{long2015fully} was the first approach to apply CNNs to semantic segmentation, laying the foundation for subsequent research. Mining the associations between contextual information in semantic segmentation is necessary. Dilated convolution \cite{chen2014semantic} was presented to enlarge the receptive field of convolution to further capture the contextual semantics. To mine multi-scale semantics, the pyramid pooling module (PPM) was proposed by \cite{zhao2017pyramid} to merge feature representations between various scales and regions. And the Astral Spatial Pyramid Pooling (ASPP) \cite{chen2017deeplab} set various dilation rates for dilated convolution to capture multi-scale contextual semantics. Meanwhile, \cite{lin2017refinenet} proposed a multiplexed tuning network to exploit visual features at different levels of low and high levels. In addition, several attention mechanism methods have been proposed to aggregate contextual semantics across long-range \cite{fu2019dual,li2019dfanet,wang2018non,li2019expectation,feng2023height}. Furthermore, to better capture details of the target (e.g., edges),  \cite{ronneberger2015u} utilized the encoder-decoder structure to merge low-level features with high-level features to obtain richer spatial detail.

Although these methods can work well on large-scale data, it's hard to achieve the desired results when handling rare or unknown data.
\subsection{Few-Shot Learning}
Recently, few-shot learning (FSL) has been proposed to tackle the above issues, aiming at recognizing novel (unknown) classes from just a few samples \cite{li2017meta}. In general, FSL methods could be divided into the following three branches: (i) Data-augmentation-based methods \cite{hariharan2017low,zhang2018metagan,wang2018low}. (ii) Transfer-learning-based methods \cite{dhillon2019baseline, chen2019closer, shen2021partial}. (iii) Meta-learning-based methods \cite{vinyals2016matching,finn2017model,ravi2017optimization,li2017meta,jamal2019task,snell2017prototypical,9501951,mao2022bidirectional}. Data-augmentation-based methods utilize unlabelled data or data synthesis methods to achieve data supplementation. And Transfer-learning-based methods pre-train models on large-scale datasets while fine-tuning models on targeted small datasets. However, both methods are limited by introducing large amounts of noise interference and generating overfitting, respectively. Meta-learning aims to learn to learn in order to facilitate rapid adaptation to new tasks based on the acquisition of existing knowledge, which is extremely well suited to Few-shot tasks that utilize a few samples to parse an unknown class. 

Therefore, the vast majority of FSL methods are designed based on the meta-learning paradigm proposed by \cite{vinyals2016matching}, which expects learning transferable meta-knowledge derived from a range of tasks (i.e., episodes) sampled from the base dataset (training dataset) to generalize to new tasks. Based on this, these meta-learning-based methods are further subdivided into the following two branches: One branch is the parametric-optimization-based method, which aims to learn the suitable parameters so that the model can be quickly adapted to new tasks \cite{finn2017model,ravi2017optimization,li2017meta,jamal2019task}. Another branch is the metric-learning-based method, which performs relevance matching and comparison to tackle few-shot problems. \cite{vinyals2016matching,snell2017prototypical} learned an embedding space and analyzed the degree of matching with a non-parametric distance function. 
% In addition, \cite{bateni2020improved,zhang2022deepemd} found more suitable distance metrics for classification.

Notably, prototype network \cite{snell2017prototypical} proposed a series of feature vectors (i.e., prototypes) to store the semantics of different categories, which could be well applied to pixel-wise segmentation tasks. Chen et al. \cite{9501951} also built on the prototype and proposed a Siamese-prototype network (SPNet) to address the limitations of Few-shot learning in remote sensing scenarios with large intra-class variance and small inter-class variance. Similarly, our work performs pixel-wise relevance matching in the form of prototypes, where prototypes store representative semantics of known classes and target classes.
\subsection{Few-Shot Segmentation}\label{Few-shot segmentation}
Few-shot segmentation (FSS) is an extension of FSL that deals with dense pixel-wise predictions with just a few samples. Shaban et al. \cite{shaban2017one} applied meta-learning to the task of few-shot segmentation for the first time and discussed the advantages of meta-learning. Existing methods generally follow their proposed meta-learning paradigm of learning transferable category semantics from support images to guide query images for segmentation. \cite{zhang2020sg} then proposed to extract prototypes (feature vectors) of the target class from the support images to guide segmentation, which was widely adopted. To better maintain the generalizability to new classes, many researchers have attempted to fix the backbone network and instead focus on implementing more efficient semantic interactions between the image pairs \cite{zhang2019canet,lang2022beyond,tian2020prior}. For example, \cite{zhang2019canet} utilized the prototypes to concatenate with all positions of the query image for comparison. Considering that a single prototype fails to represent the entire target feature, \cite{lang2022beyond} extracted multiple prototypes with different roles from the support image for guidance. 

Recently some researchers have extended the Few-shot segmentation task to remote sensing\cite{jiang2022few,chen2022semi,yao2021scale,wang2021dmml,10152484}. Jiang et al. \cite{jiang2022few} applied metric learning to Few-shot segmentation in remote sensing, thus solving the problem of insufficient labeled data in remote sensing scenarios. Chen et al. \cite{chen2022semi} employed a novel Few-shot segmentation framework to better distinguish features by mining the latent novel classes in the contexts via self-supervised learning. Also, Lang et al.\cite{10152484} proposed to mine the semantics between the image pairs and between classes, in consideration of the characteristics of remote sensing scenes with large intra-class differences and low foreground-background contrast. Different from them, we take inspiration from the human process of learning knowledge and rethink the task of Few-shot segmentation, arguing that not only do we need to learn from others but we also need to focus on ourselves. i.e., in addition to mining the common semantics between the support-query image pair, one can also mine the specific semantics of the query image itself.

Furthermore, another reason that limits the performance of Few-shot segmentation is the bias toward known classes. \cite{lang2022learning} proposed to set up an additional learner to learn non-target regions, thus aiding the main learner for target segmentation. Meanwhile, \cite{cheng2022holistic} proposed a two-stage method to learn knowledge from the base class to facilitate segmentation. However, training an additional learner or performing two-stage is time-consuming and inefficient. Instead, we propose continuously learning the semantics of known classes during the regular training phase and assisting in segmenting targets. Significantly, our one-stage method is nearly parameter-free, and no extra training is required.

\section{Problem Definition}\label{Problem Definition}
Unlike traditional semantic segmentation, which can only segment targets of known classes (training classes), FSS aims to segment unknown-class targets with just a few annotated samples without additional training. Notably, FSS performs single-class segmentation, i.e., only novel class (unknown class) is segmented while other classes are considered as background. Current FSS methods usually follow the meta-learning paradigm for training models (i.e., episodic training), which can learn a generic segmentation capability to generalize unknown classes in each episodic. Specifically, the whole dataset is usually divided into two subsets, a training set \emph{D}$_{train}$ with known classes \emph{C}$_{known}$ and a testing set \emph{D}$_{test}$ with unknown classes \emph{C}$_{unknown}$. Note that the categories of these two sets are disjoint (i.e., $ C_{known} \cap C_{unknown}= \emptyset $). The FSS models learn transferable latent knowledge on \emph{D}$_{train}$ with a sufficient amount of annotated samples and demonstrate satisfactory generalization to \emph{D}$_{test}$ with a small amount of annotated samples. In particular, both sets \emph{D}$_{train}$ and \emph{D}$_{test}$ consist of a number of episodes, each containing a support set $ S =\left \{ \left ( X_i^s, M_i^s\right )  \right \}_{i=1}^{K} $  and a query set $ Q =\left \{ \left ( X^q, M^q\right )  \right \} $, where $X^{*}_{i}$ and $M^{*}_{i}$ denote the original image and the binary mask corresponding to a particular class $c$ (both $X^{s}$ and $X^{q}$ contain the category $c$), respectively. It is noteworthy that $K$ represents the number of support images given; following previous work we only explore the FSS performance for $K$= 1 and $K$= 5 (i.e., 1-shot and 5-shot). During each training episode, the model mines the information in the support set \emph{S} to segment the relevant regions of category $c$ in the query image $X^{q}$. After the training is completed, the model evaluates the performance in \emph{D}$_{test}$, at which point the parameters of the model are not optimized. Significantly, the mask $M^{q}$ in the query set \emph{Q} is provided only during the training phase.
\section{Proposed Method}\label{Proposed Method}
\subsection{Method Overview}
\begin{figure*}[t]
	\setlength{\abovecaptionskip}{2pt}
	\centering
	\includegraphics[width=1.0\linewidth]{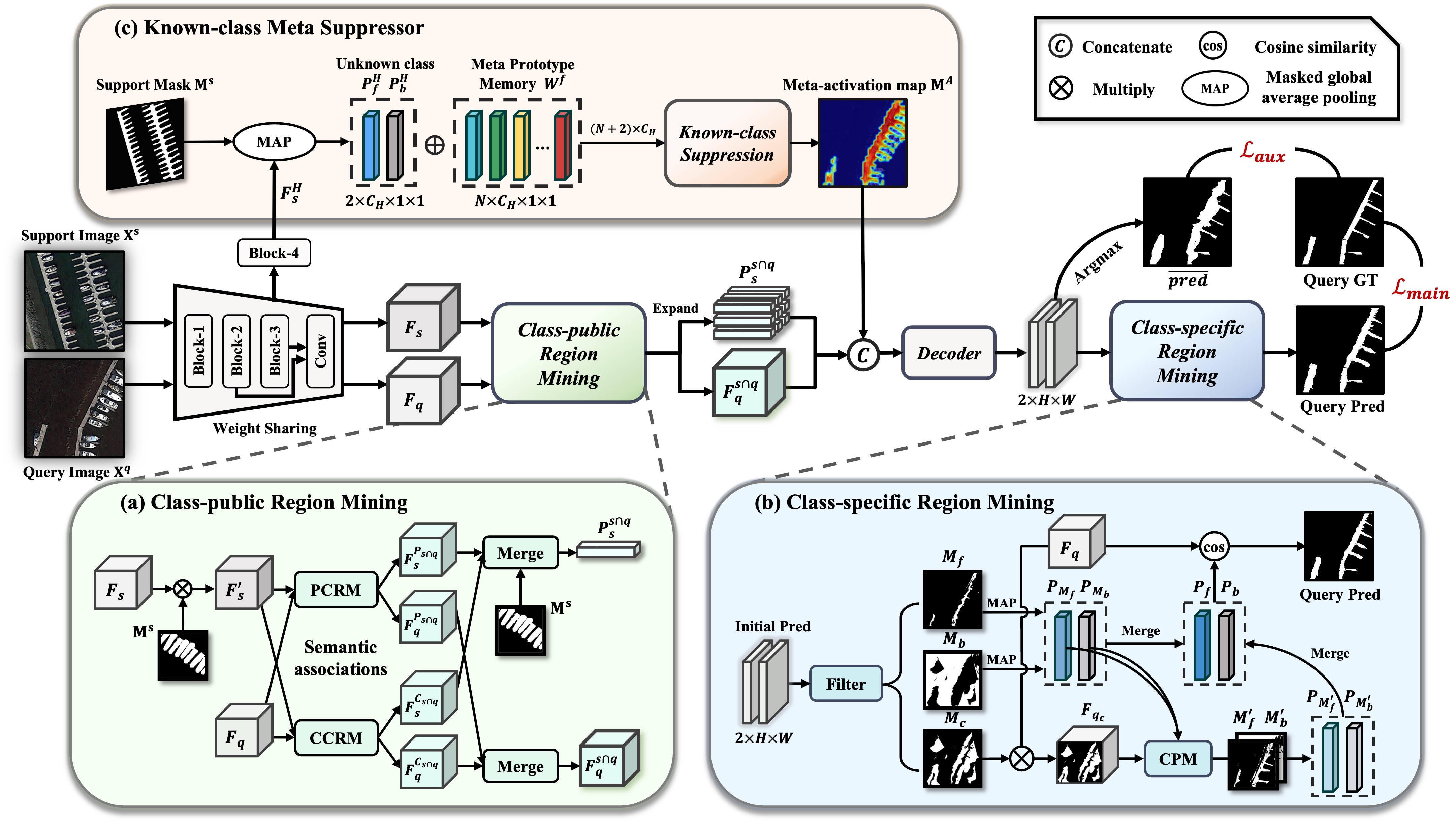}
	\caption{The overall framework of the proposed network DMNet, consists of three significant modules, including Class-public Region Mining (CPRM), Class-specific Region Mining (CSRM), and Known-class Meta Suppressor (KMS). PCRM represents the Position-based Class-public Region Mining Module which captures spatial correlation based on category semantics in support features and query features and CCRM represents the Channel-based Class-public Region Mining Module which is similar to PCRM for capturing channel correlation. Filter represents the Feature Filtering Module, and CPM represents the Confusion-region Prototype Module.}
	\label{fig:2}
\end{figure*}
As mentioned above, previous FSS methods are susceptible to intra-class variation and suffer from the collapse of generalization in remote sensing scenes. To this end, we propose DMNet, a cross-image mining and self-information mining network that contains Class-public Region Mining (CPRM), Class-specific Region Mining (CSRM), and Known-class Meta Suppressor (KMS) three important components as shown in Fig.\ref{fig:2}, aiming at addressing these two problems.

Precisely, given a support-query image pair $X^s$ and $X^q$, we follow the previous approaches and leverage the pre-trained CNN backbone to extract features to obtain both the support feature $F^s$ and the query feature $F^q$. The proposed CPRM module then activates the class-public semantics between $F^s$ and $F^q$ and weakens irrelevant semantics, which can somewhat mitigate intra-class feature differences and suppress useless feature pollution from MAP operations. In particular, we utilize Position-based Class-public Region Mining (PCRM) and Channel-based Class-public Region Mining (CCRM) to model the semantic association of target categories between features in the position and channel dimensions, respectively. Thus, through the CPRM module, the activated common query feature $F_q^{s \cap q}$ and the support prototype $P_s^{s \cap q}$ are derived. Next, we expand  $P_s^{s \cap q}$ to match the size of the query feature $F_q^{s \cap q}$ and feed them together into the decoder to get the initial prediction $\bm{y}^q$. 

As the category semantics mined from the support image cannot be fully matched to the query image, the proposed CSRM module mines the category semantics of the query image itself to guide its own segmentation. Concretely, the module filters and purifies the prediction result $\bm{y}^q$ by a filter to derive the latent category semantics $P_{M_{f}}$ and $P_{M_{b}}$ of the query image. Then the Confusion-region Prototype Module (CPM) exploits these semantics to activate other target regions $M_{f}^{'}$ and $M_{b}^{'}$ in a continuously guided manner. More importantly, the CPRM and CSRM modules complement each other and work together, with the former providing sufficient quality initial predictions and the latter capturing class-specific semantics that the former can't.

Besides, to alleviate the collapse of generalization to unknown classes (over-fitting to irrelevant known classes) in remote sensing scenarios, we design a unique meta-training paradigm and a corresponding KMS module that introduces an additional branch during the training phase to continuously capture the representative semantics of known classes utilizing a meta-memory W$^{f}$. The Known-class Suppression Module then utilizes the captured known class prototypes W$^{f}$ and the target prototypes $P_{f}^{H} (P_{b}^{H})$ to jointly suppress the activation of known classes in the query image. A meta-activation map M$^{A}$ with the known classes suppressed is thus obtained, while it is also fed to the decoder for subsequent operations.
\subsection{Class-public Region Mining} \label{sec:Class-public Region Mining (CPRM)}
As mentioned in Section \ref{Indroduction}, blindly migrating the class semantics of support images to guide query image segmentation will inevitably introduce irrelevant semantics. We expect to capture similar `fuselage' semantics while filtering out irrelevant `wing' semantics. Inspired by the fact that self-attention \cite{fu2019dual,wang2018non} builds semantic associations between pixel pairs well, capturing common category semantics between the image pair with cross-attention is a natural choice. Thus, we propose the Class-public Region Mining (CPRM) module, which aims to capture bidirectional semantic associations between the support-query image pair, thereby suppressing semantics in the support image that are not useful for the query image and locating the target region of the query image.

In detail, given the support feature $\bm{F}_s$ and query feature $\bm{F}^q \in \mathbb{R}{^{C \times H \times W}}$, we perform a dot product operation on support feature $\bm{F}_s$ with its corresponding mask $\bm{M}^s$ to activate only the features of the target category, which can be formulated as $\bm{F}_{s}^{'} = \bm{F}_{s} \odot{\bm{M}^s}$, where $\odot{}$ denotes the dot product operation. Remarkably, the subsequently mentioned support features all represent the part of the support features that only activate the target class region.

\begin{figure}[t]
	\setlength{\abovecaptionskip}{2pt}
	\centering
	\includegraphics[width=1.0\linewidth]{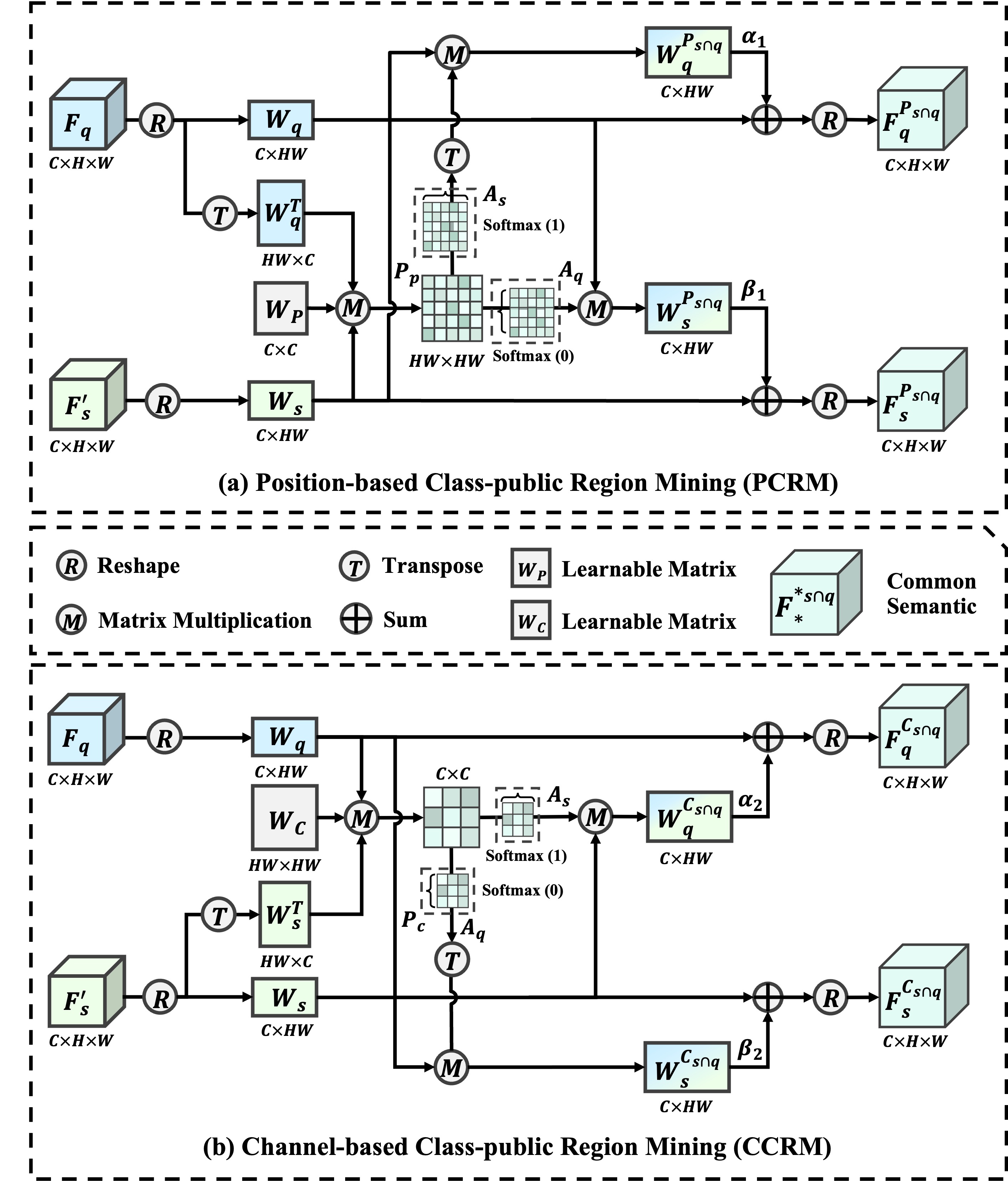}
	\caption{The specific structure of PCRM and CCRM modules. Here $\bm{F}^{'}_{s}$ represents the support feature that activates only the target category. Notably, softmax(0) and softmax(1) represent the normalization by columns and rows for convenience, respectively.}
	\label{fig:3}
\end{figure}

Then, as shown in Fig.\ref{fig:3}, the PCRM and CCRM modules are proposed to capture bidirectional semantic associations in the position dimension and the channel dimension, respectively.

\subsubsection{Position-based Class-public Region Mining}\label{sec:Position-based Class-public Region Mining} 

We reshape $\bm{F}_{q}$ and $\bm{F}_{s}^{'}$ into two-dimensional features $\bm{W}_q$ and ${\bm{W}_s \in \mathbb{R}{^{C \times HW}}}$ while transposing $\bm{W}_q$ into ${\bm{W}_q^{T} \in \mathbb{R}{^{HW \times C}}}$. Then we compute the affinity matrix $\bm{L}_P$ between $\bm{W}_q^{T}$ and $\bm{W}_s$, which can be formulated as:
\begin{equation}\label{equation:1}
\setlength{\abovecaptionskip}{1pt}
\setlength{\belowcaptionskip}{1pt}
\bm{L}_P = \bm{W}_q^{T}\bm{W_P}\bm{W}_s 
\end{equation}
where ${\bm{W_P} \in \mathbb{R}{^{C \times C}}}$ is used to balance the scales of the features and its parameters are learnable. The $\bm{L}_P(i, j)$ in the affinity matrix $\bm{L}_P$ represent the semantic similarity between the $i^{th}$ position in $\bm{W}_q^{T}$ and the $j^{th}$ position in $\bm{W}_s$. Since the $\bm{F}^{'}_{s}$ only activates the region of the target category, a higher similarity score indicates that the pixel at $i^{th}$ position in the query image is likelier to be the target category. 
 
Then we normalize the affinity matrix $\bm{L}_P$ by columns to generate attention maps $\bm{A}_q$ for each position in $\bm{W}_s$ with respect to $\bm{W}_q$, and by rows to generate attention maps $\bm{A}_s$ for each position in $\bm{W}_q$ with respect to $\bm{W}_s$.

Next, we obtain the position-based semantic effects of support feature (query feature) on query feature (support feature) based on the weighted aggregation of $\bm{W}_s (\bm{W}_q)$ and $\bm{A}_s (\bm{A}_q)$.

Finally, we design a learnable weight to fuse the common semantics with the features, which can be formulated as:
\begin{align}
\setlength{\abovecaptionskip}{1pt}
\setlength{\belowcaptionskip}{1pt}
 \bm{F}_q^{P_{s \cap q}} &= {{\mathcal{F}}_{{reshape}}}(\alpha_1 \bm{W}_s\bm{A}_s+ \lambda \bm{W}_q)  \\
 \bm{F}_s^{P_{s \cap q}} &= {{\mathcal{F}}_{{reshape}}}(\beta_1 \bm{W}_q\bm{A}_q+ \lambda \bm{W}_s) 
\end{align}
where $\alpha_1$ and $\beta_1$ represent the learnable fusion weights, their initial values are set to 0.5. $\lambda$ represents a fixed fusion weight, which is set to 0.5. Here ${{\mathcal{F}}_{{reshape}}}$ reshapes the size of input sensor to $C \times H \times W$.

Therefore, we build a bidirectional semantic association between support-query image pair to focus on the common semantics (i.e., the target class region), thus alleviating the gap of intra-class variance.

\subsubsection{Channel-based Class-public Region Mining}\label{sec:Channel-based Class-public Region Module}
Meanwhile, the CCRM module also builds bidirectional semantic associations in the channel dimension to better focus on the public semantics of the target, which is similar to PCRM. Thus, we can obtain the channel-based class-public semantics $\bm{F}_q^{C_{s \cap q}}$ and $\bm{F}_s^{C_{s \cap q}}$.

\subsubsection{Feature Aggregation}\label{sec:Feature Aggregation} 
After mining the position-based and channel-based class-public semantics, we perform feature aggregation to obtain the final semantic features $\bm{F}_q^{{s \cap q}}$ and $\bm{F}_s^{{s \cap q}} \in \mathbb{R}{^{C \times H \times W}}$. Here, we perform MAP operation on activated support features $\bm{F}_s^{{s \cap q}}$ to obtain a more representative support prototype $\bm{P}_s^{{s \cap q}} \in \mathbb{R}{^{C \times 1 \times 1}}$, which can be formulated as:
\begin{align}
\setlength{\abovecaptionskip}{1pt}
\setlength{\belowcaptionskip}{1pt}
\bm{F}_q^{{s \cap q}} &= {{\mathcal{F}}_{1\times 1}}(\bm{F}_q^{P_{s \cap q}}+\bm{F}_q^{C_{s \cap q}})  \\
\bm{F}_s^{{s \cap q}} &= {{\mathcal{F}}_{1\times 1}}(\bm{F}_s^{P_{s \cap q}}+\bm{F}_s^{C_{s \cap q}})  \\
\bm{P}_s^{{s \cap q}} &= {{\mathcal{F}}_{MAP}}(\bm{F}_s^{{s \cap q}} \odot \bm{M}^s) 
\end{align}
where ${{\mathcal{F}}_{1\times 1}}$ indicates a $1 \times 1$ convolution, ${{\mathcal{F}}_{MAP}}$ indicates the masked average pooling operation and $\odot$ indicates dot product operations.

Through the CPRM module, the query feature $\bm{F}_q^{{s \cap q}}$ activates target category regions that share common attention with the support feature while suppressing other non-target regions. For the support feature $\bm{F}_s^{{s \cap q}}$, the category semantics representing the targets of the support image themselves are suppressed, while the more generic public category semantics are retained. Therefore, the more precise target regions in query feature $\bm{F}_q^{{s \cap q}}$ are activated under the guidance of $\bm{P}_s^{{s \cap q}}$.

Meanwhile, to more accurately locate the target regions, we average the rows of the affinity matrix $\bm{P}_p$ in the PCRM module in anticipation of obtaining the positional activation map $\bm{M}^p$ of the query feature.

Thus, after the CPRM module, the initial prediction result $\bm{y}^q \in \mathbb{R}{^{2 \times H \times W}}$ is derived through a simple decoder network as follows:
\begin{align}\label{equation:12}
\setlength{\abovecaptionskip}{1pt}
\setlength{\belowcaptionskip}{1pt}
\bm{y}^q = {{\mathcal{F}}_{Dec}}(\bm{F}_q^{s \cap q} \oplus \bm{P}_s^{s \cap q} \oplus \bm{M}^p )
\end{align}
where $\bm{P}_s^{s \cap q}$ is expanded to the same shape as $\bm{F}_q^{s \cap q}$ and $\oplus$ represents the concatenation operation along channel dimension. ${{\mathcal{F}}_{Dec}}$ represents the decoder network, which consists of several $3 \times 3$, $1 \times 1$ convolutional layers and ASPP \cite{chen2017deeplab}. 

\subsection{Class-specific Region Mining}\label{sec: Class-specific Region Mining (CSRM)}
Many FSS methods end at this step, but bridging the gap of intra-class variance by mining only the semantic associations between support and query features is still hard. Therefore, it may be good thinking to focus more on mining the query image itself. We argue that even if there is some false activation in the initial prediction result $\bm{y}^q$, there must be latent category semantics specific to the query image itself that can be captured and exploited. Based on this, we design the Class-specific Region Mining (CSRM) module.

Specifically, given the initial predicted results $\bm{y}^q \in \mathbb{R}{^{2 \times H \times W}}$ which contains a foreground prediction $\bm{y}^q_f \in \mathbb{R}{^{1 \times H \times W}}$ and a background prediction $\bm{y}^q_b \in \mathbb{R}{^{1 \times H \times W}}$. We perform a semantic filter on $\bm{y}^q$: (i) Regard regions with foreground prediction scores above a certain threshold as foreground regions $\bm{M}_f$ containing richer target semantics of the query image-self; (ii) Regard regions with background prediction scores above a certain threshold as background regions $\bm{M}_b$ containing richer background semantics of the query image-self; (iii) We also consider the other regions (i.e., regions with low scores in both the foreground and background predictions) as confusion regions $\bm{M}_c$ that are prone to confusion and false activation:
\begin{align}
\setlength{\abovecaptionskip}{1pt}
\setlength{\belowcaptionskip}{1pt}
\bm{M}_f &= {{\mathcal{F}}_{Indicator}} \left [ \bm{y}^q_f \ge \mu_1  \right ]  \label{equation:13}\\
\bm{M}_b &= {{\mathcal{F}}_{Indicator}} \left [ \bm{y}^q_b \ge  \mu_2  \right ]  \label{equation:14}\\
\bm{M}_c &= {{\mathcal{F}}_{Indicator}}\left [ \left ( \bm{y}^q_f <\mu_1 \right )\cap  \left ( \bm{y}^q_b <\mu_2 \right ) \right ]
\end{align}
where $\bm{y}^q_f = \mathrm{softmax}(\bm{y}^q)[1,:]$ and $\bm{y}^q_b = \mathrm{softmax}(\bm{y}^q)[0,:]$. ${{\mathcal{F}}_{Indicator}}$ denotes the indicator function which converts the predicted probability result into a prediction mask. The thresholds $\mu_1$ and $\mu_2$ indicate the filtering degree of the predicted results $\bm{y}^q$, set to 0.7 and 0.6, respectively.

Thus, the foreground prototype $\bm{P}_{M_f}$ and background prototype $\bm{P}_{M_b}$ of the target category are mined from $\bm{M}_f$ and $\bm{M}_b$ by the MAP operation, respectively. We argue that it is more effective to utilize these prototypes, containing rich semantics specific to the query image itself, to guide its own segmentation.
\begin{figure}[t]
	\setlength{\abovecaptionskip}{2pt}
	\centering
	\includegraphics[width=1.0\linewidth]{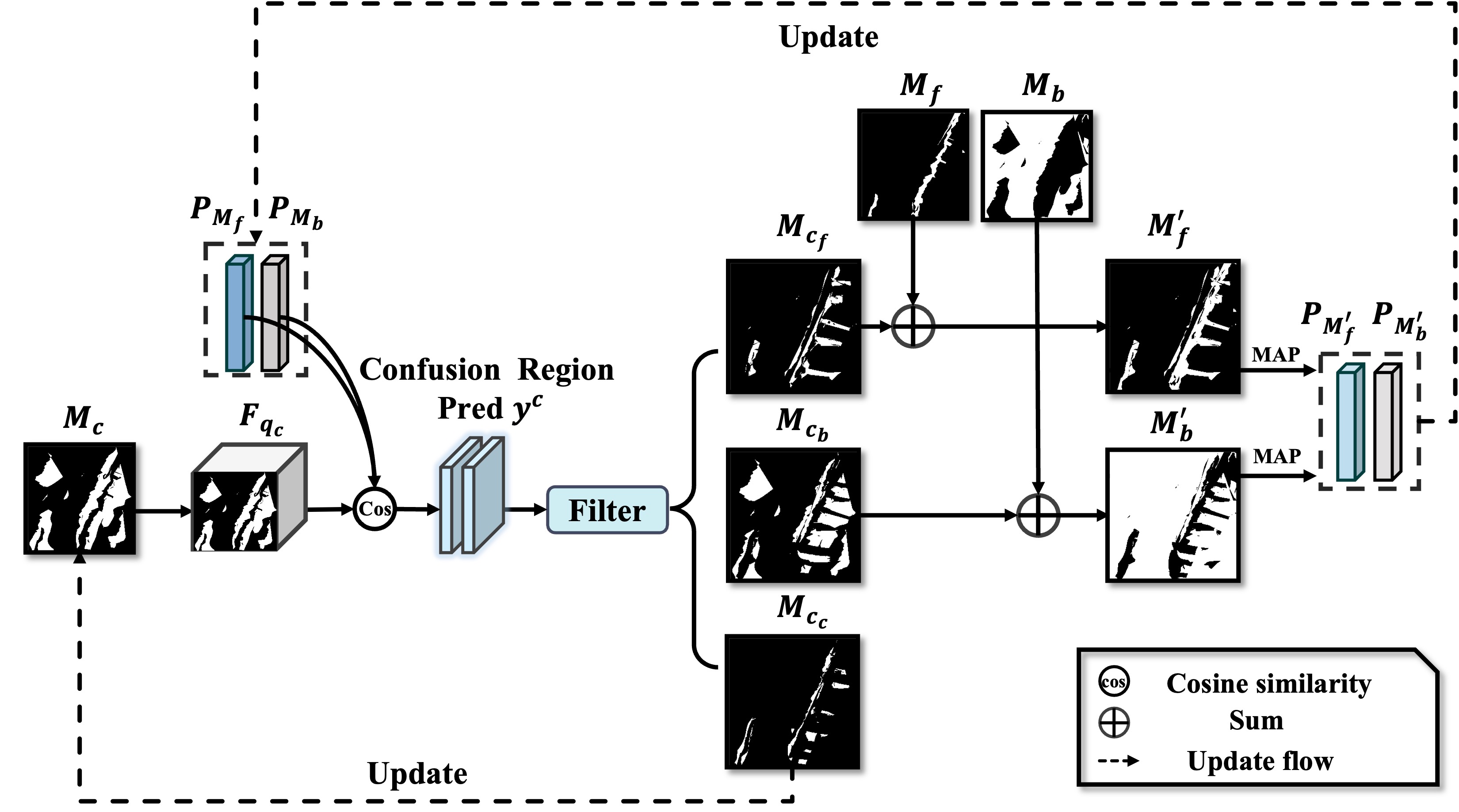}
	\caption{The specific structure of Confusion-region Prototype Module (CPM). Note that the dashed arrows represent the flow of iterative mining of the confusion region. }
	\label{fig:4}
\end{figure}

In addition to this, we expect that there are also rich latent semantics of the target category in the confusion region $\bm{M}_c$ and hence design a Confusion-region Prototype Module (CPM) to capture. Specifically, as shown in Fig.\ref{fig:4}, we utilize the mined $\bm{P}_{M_f}$ and $\bm{P}_{M_b}$ to activate the foreground and background regions in the confusion regions $\bm{M}_c$ in an associated manner to generate the prediction $\bm{y}^c \in \mathbb{R}{^{2 \times H \times W}}$:
\begin{align}
\setlength{\abovecaptionskip}{1pt}
\setlength{\belowcaptionskip}{1pt}
\bm{y}^c = \mathrm{softmax}({{\mathcal{F}}_{cosine}}(\bm{F_{q_c}},\bm{P}_{M_f},\bm{P}_{M_b}))  
\end{align}
where $\bm{F}_{q_c} = \bm{F}_q \odot \bm{M}_c$ denotes the feature of confusion regions $\bm{M}_c$. Here we define a function ${{\mathcal{F}}_{cosine}}$ to calculate the similarity between features and different prototypes to obtain the prediction, i.e., ${{\mathcal{F}}_{cosine}}(\bm{F_{q_c}},\bm{P}_{M_f},\bm{P}_{M_b},\cdots)=\mathrm{cosine}(\bm{F_{q_c}},\bm{P}_{M_f})\oplus \mathrm{cosine}(\bm{F_{q_c}},\bm{P}_{M_b})\oplus\mathrm{cosine}(\bm{F_{q_c}},\cdots)$.

After that, we filter and purify the prediction $\bm{y}^c$ to obtain the foreground region $\bm{M}_{c_f}$, background region $\bm{M}_{c_b}$, and confusion region $\bm{M}_{c_c}$ in the confusion region $\bm{M}_c$. 
We expect a more representative foreground prototype $\bm{P}_{M_f^{'}}$ and a background prototype $\bm{P}_{M_b^{'}}$ after fusing the foreground region $\bm{M}_{c_f}$ and the background region $\bm{M}_{c_b}$ of the confusion region $\bm{M}_c$, which can be formulated as:
\begin{align} 
\setlength{\abovecaptionskip}{1pt}
\setlength{\belowcaptionskip}{1pt}
\bm{P}_{M_f^{'}} &= {{\mathcal{F}}_{MAP}}(\bm{F}_q \odot(\bm{M}_f + \bm{M}_{c_f}))   \\
\bm{P}_{M_b^{'}} &= {{\mathcal{F}}_{MAP}}(\bm{F}_q \odot(\bm{M}_b + \bm{M}_{c_b})) 
\end{align}
where $\bm{M}_{c_f}$ and $\bm{M}_{c_b}$ are derived from the execution of Equation \ref{equation:13} and Equation \ref{equation:14} by $\bm{y}^c$.

It is worth noting that we repeat the above filtering process several times with the new confusion regions to obtain the final foreground (background) prototype, with the aim of sufficiently mining the semantics in the confusion regions to capture more category semantics from the query image itself. In our experiments, we set 3 iterations. And in each filtering, our confidence thresholds $\mu_1$ and $\mu_2$ are each gradually reduced by 0.05 and 0.02 to derive more category semantics. 

Subsequently, we weight the merging of prototypes $\bm{P}_{M_f^{'}} \left ( \bm{P}_{M_b^{'}} \right)$ and $\bm{P}_{M_f} \left ( \bm{P}_{M_b} \right)$ to avoid the reduction in prototype expressiveness caused by errors in CPM filtering:
\begin{align}
\setlength{\abovecaptionskip}{1pt}
\setlength{\belowcaptionskip}{1pt}
\bm{P}_f = \gamma_1\bm{P}_{M_f^{'}} + \gamma_2\bm{P}_{M_f}, \bm{P}_b = \gamma_1\bm{P}_{M_b^{'}}+ \gamma_2\bm{P}_{M_b} 
\end{align}
where $\gamma_1$ and $\gamma_2$ are the integration weights and we set $\gamma_1=0.9$ and $\gamma_2 = 0.1$ in our experiments. 

We then compute the cosine similarity between the augmented prototypes $\bm{P}_f \left ( \bm{P}_b \right)$ and the query feature $\bm{F_q}$ to derive the final segmentation prediction: 
\begin{align}  \label{equation:25}
\setlength{\abovecaptionskip}{1pt}
\setlength{\belowcaptionskip}{1pt}
\bm{y}^{Final} = \mathrm{softmax}({{\mathcal{F}}_{cosine}}(\bm{F_{q}},\bm{P}_f,\bm{P}_b)) 
\end{align}
where ${{\mathcal{F}}_{cosine}}$ represents the process of computing similarity between the query feature $\bm{F}_q$ and $\bm{P}_f$ and $\bm{P}_b$ respectively to generate the prediction result.

\subsection{Known-class Meta Suppressor}\label{sec:Known-class Meta Suppressor (KMS)}
The co-existence of multiple categories in remote sensing scenarios can worsen the generalization to unknown categories, where irrelevant known classes will be activated incorrectly. We argue that since it is inevitable to introduce the knowledge of known classes, it is simply a matter of capturing representative semantics of known classes during the training phase and suppressing the activation of known classes in the query image with their help. Thus, we propose a unique meta-training paradigm and corresponding Known-class Meta Suppressor (KMS) module.

Since the model samples a series of episodes during the training phase to mimic few-shot scenarios with unknown classes, where these episodes are for known classes and contain no unknown classes, the KMS module is set up differently during the training and testing phases.
\subsubsection{KMS during the training phase}\label{sec: KMS during the training phase}
\begin{figure*}[t]
	\setlength{\abovecaptionskip}{2pt}
	\centering
	\includegraphics[width=1.0\linewidth]{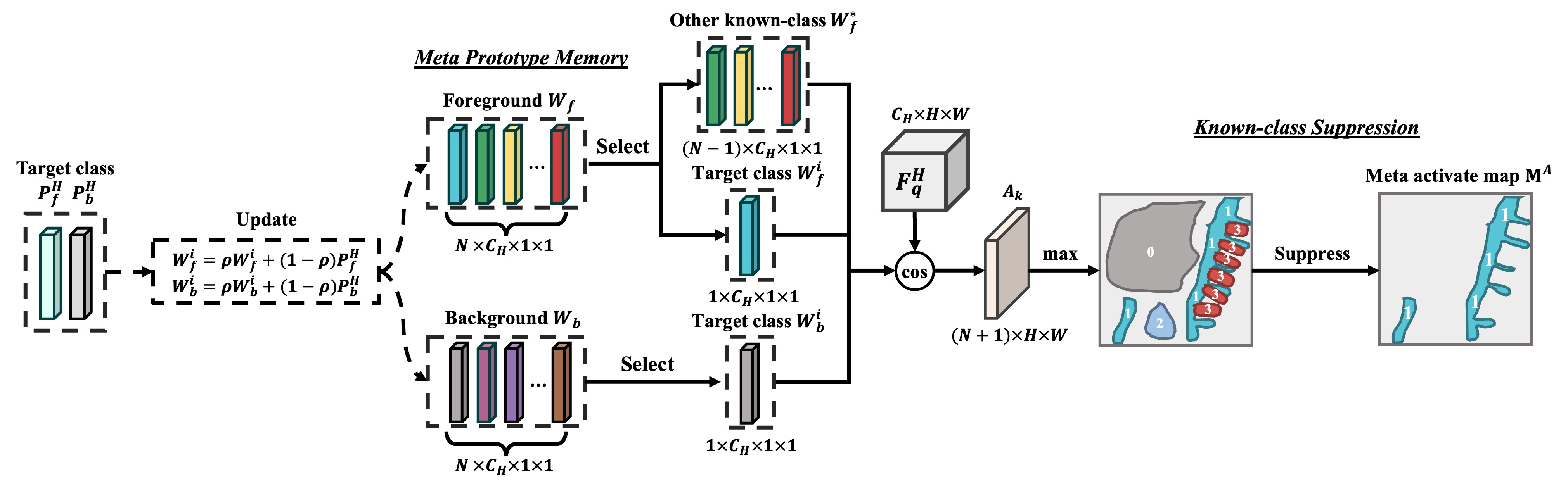}
	\caption{The specific structure of the Known-class Meta Suppressor (KMS) module during the training phase. The \textbf{`target class'} refers to the target class under the current episode, i.e., the class contained in both the support image and the query image. $\bm{N}$ represents the number of known classes during the training phase. Note that $\bm{N}$ is set to 10 for the iSAID, 4 for the LoveDA, and 15 for the PASCAL-5$^i$ in the experiments. And $\bm{C}_k$ represents the number of channels of the high-level feature. The numbers `$1$' and `$0$' in $\bm{M}^A$ denote the foreground and background regions of the target class, respectively, and the other numbers denote the foreground regions of other known classes (i.e., irrelevant regions). }
	\label{fig:5}
\end{figure*}

To obtain the representative semantics of known classes, a new meta-training paradigm is designed, where an additional branch is introduced during the training phase to learn the known-classes semantics continuously. Specifically, we propose a Meta Prototype Memory that mines and stores the prototype semantics of the current target class from the support image in each episode. Thus, the prototypes of all training classes are stored in the meta-memory after several episodes, where $\bm{N}$ denotes the number of all training classes. Notably, we choose to mine the high-level features since the semantics in the high-level features are more class-specific than the mid-level features. In addition, considering that there is some similarity and generality in the background of the same category, we likewise mine the general background semantics of known classes to distinguish them better; thus the memory consists of two parts: the foreground prototype memory $\bm{W}_f$ and the background prototype memory $\bm{W}_b$.

In particular, as shown in Fig.\ref{fig:5}, in each episode the Meta Prototype Memory utilizes foreground $\bm{P}_f^H$ and background $\bm{P}_b^H$ prototypes from high-level support features to update the representative prototypes $\bm{W}_f^i$ and $\bm{W}_b^i$ of the current class in the memory, which can be formulated as:
\begin{align}
\setlength{\abovecaptionskip}{1pt}
\setlength{\belowcaptionskip}{1pt}
\bm{W}_f^i=\rho\bm{W}_f^i+(1-\rho)\bm{P}_f^H,\bm{W}_b^i=\rho\bm{W}_b^i+(1-\rho)\bm{P}_b^H 
\end{align}
where $\bm{P}_f^H = {{\mathcal{F}}_{MAP}}(\bm{F}_s^H \odot \bm{M}^s)$ and $\bm{P}_b^H = {{\mathcal{F}}_{MAP}}(\bm{F}_s^H \odot (1-\bm{M}^s))$. After several episodes, the Meta memory is continuously updated to obtain representative prototypes for all training categories (i.e., $\bm{N}$ representative prototypes). In the experiment, $\rho$ is set to 0.5.

It is easy to see that the more times the prototypes are updated in the Meta Prototype Memory, the more representative they are (more prototypes are used to update). Instead suppressing the activation of non-target regions in the query image with the known class prototypes that are updated only a few times will degrade the segmentation performance. Therefore, we propose that the parameters of the model only start to be updated after a certain number of training iterations, i.e., only the known-class prototypes in the memory are updated during the first few epochs while the parameters of the model are fixed. It is not until these prototypes have become more representative that the parameters start to be updated.

We then select from memory the representative prototypes of the foreground $\bm{W}_f^i \in \mathbb{R}{^{\bm{C}_H \times 1 \times 1}}$ and background $\bm{W}_b^i \in \mathbb{R}{^{\bm{C}_H \times 1 \times 1}}$ of the current target class, and prototypes $\bm{W}_f^* \in \mathbb{R}{^{(\bm{N}-1) \times \bm{C}_H \times 1 \times 1}} $ of the foreground of remaining known classes. Thus, $(\bm{N}+1)$ representative semantic prototypes compute the similarity with the high-level query feature $\bm{F}_q^H$ to obtain $(\bm{N}+1)$ prediction maps $\bm{A}_k$:
\begin{equation}\label{equation:23}
\setlength{\abovecaptionskip}{1pt}
\setlength{\belowcaptionskip}{1pt}
\begin{split}
   \bm{A}_k &= \bm{A}_{0} \oplus \bm{A}_{1} \oplus \bm{A}_{2} \oplus \cdots \oplus \bm{A}_{N} \\
     &={{\mathcal{F}}_{cosine}}(\bm{F}_q^H,\bm{W}_f^i,\bm{W}_b^i,\bm{W}_f^* )
\end{split}
\end{equation}
where $k \in \left [0, \bm{N} \right]$. $\bm{A}_0$ denotes the prediction result of computing similarity with background semantics $\bm{W}_b^i$ of the current target class, $\bm{A}_1$ denotes the prediction result of computing similarity with foreground semantics $\bm{W}_f^i$, while $k$ taking other values denotes the prediction result of other known classes (i.e., non-target classes).

The maximum value and corresponding index value are then calculated for each position in the query feature. Finally, we suppress the regions activated by the prototypes of non-target classes $\bm{W}_f^*$ and the background prototype of current target class $\bm{W}_b^i$ to mitigate the effects of the other known classes:   
\begin{align}
\setlength{\abovecaptionskip}{1pt}
\setlength{\belowcaptionskip}{1pt}
&K^{(m,n)} = { \underset {k} { \operatorname {arg\,max}}}(\bm{A}_k^{(m,n)}) \\
&\bm{M}^A = {{\mathcal{F}}_{Indicator}}\left [K^{(m,n)}=1 \right ]\bm{A}_{1} 
\end{align}
where $K^{(m,n)}$ indicates the index value of the representative prototype which obtains maximum similarity score (i.e., is activated) at the position (m,n) of the query feature. ${{\mathcal{F}}_{Indicator}}\left [K^{(m,n)}=1 \right ]$ denotes the activated foreground regions of the current target class.

Thus we obtain a meta-activation map $\bm{M}^A$ that suppresses other known classes and background regions, and we feed it together into the decoder, at which point Equation \ref{equation:12} becomes:
\begin{align}
\setlength{\abovecaptionskip}{1pt}
\setlength{\belowcaptionskip}{1pt}
\bm{y}^p = {{\mathcal{F}}_{Dec}}(\bm{F}_q^{s \cap q} \oplus \bm{P}_s^{s \cap q} \oplus \bm{M}^p \oplus \bm{M}^A)
\end{align}
\subsubsection{KMS during the testing phase}\label{sec:KMS during the testing phase}

As shown in Fig.\ref{fig:2}, we have not to update the memory only to utilize the representative prototypes of the known classes in the memory during the testing phase. It is worth noting that all known classes can be considered interference objects during the testing phase, and therefore all known classes appearing in the query image need to be suppressed. Since there is some similarity in the backgrounds of the same category, i.e. are category representative, the background prototype of the support image can also be employed to suppress the activation of the background (non-target) regions in the query image, at which point Equation \ref{equation:23} will be updated to:
\begin{equation}
\setlength{\abovecaptionskip}{1pt}
\setlength{\belowcaptionskip}{1pt}
\begin{split}
   \bm{A}_k &= \bm{A}_{0} \oplus \bm{A}_{1} \oplus \cdots \oplus \bm{A}_{N} \oplus \bm{A}_{N+1} \\
     &= {{\mathcal{F}}_{cosine}}(\bm{F}_q^H,\bm{P}_f^H,\bm{P}_b^H,\bm{W}_f )
\end{split}
\end{equation}
where $\bm{W}_f \in \mathbb{R}{^{\bm{N} \times \bm{C}_H \times 1 \times 1}} $ contains all known class representative prototypes and in this case $k \in [0, (\bm{N}+1)]$. Here $\bm{A}_0$ and $\bm{A}_1$ denote the similarity scores for the foreground and background of the current target class, respectively, while $\bm{A}_k$ $(k \in [2, \bm{N}+1])$ denotes the similarity scores of all known classes.

We then exploit the Known-class Suppression Module to suppress the activation of all known classes in the query image. Thus we can obtain a corresponding meta-activation map $\bm{M}^A$ for the testing phase, which is also fed into the decoder for subsequent operations.
\subsection{Loss Function}\label{sec:Loss Function}
To better optimize the parameters of the model during the training phase, we employ two binary cross entropy (BCE) losses $\mathcal{L}_{aux}$ and $\mathcal{L}_{main}$ to supervise the prediction results $\bm{y}^q$ and $\bm{y}^{Final}$, which constitute the overall target segmentation loss $\mathcal{L}$:
\begin{equation}
\setlength{\abovecaptionskip}{1pt}
\setlength{\belowcaptionskip}{1pt}
\begin{split}
   \mathcal{L} &= \mathcal{L}_{main} + \eta \mathcal{L}_{aux} \\
     &= \mathrm{\bm{BCE}}(\bm{y}^{Final}, \bm{M}^q) + \eta \mathrm{\bm{BCE}}(\bm{y}^q, \bm{M}^q)
\end{split}
\end{equation}
where $\eta$ is employed to balance the contributions of the two losses, $\mathcal{L}_{main}$ and $\mathcal{L}_{aux}$.
\subsection{K-shot Setting}\label{sec:K-shot Setting}
When one-shot is extended to the K-shot setting, FSS models generally average support prototypes derived from multiple support images and employ the averaged prototypes to guide the query image for segmentation. This prototype average approach simply assumes that each support image contributes equally to the query image \cite{tian2020prior}. However, due to the inconsistent feature differences between the different support images and the query image, this prototype average method is sub-optimal. Thus, We propose a new appearance similarity-based Reweighted-Fusion mechanism to handle the contributions of different support images.

In particular, we guide the query image using $\mathrm{\bm{K}}$ different support images to obtain $\mathrm{\bm{K}}$ segmentation branches respectively. As mentioned in Section \ref{sec:Class-public Region Mining (CPRM)}, through the CPRM module each branch obtains the affinity matrix $\bm{L}_P^j$, where $j \in \left [1,\mathrm{\bm{K}} \right]$. We believe that the obtained affinity matrix can represent the appearance similarity between the support image and the query image, so we efficiently compress the affinity matrix as the appearance factor $\varphi_j$, which can be formulated as:
\begin{align}
\setlength{\abovecaptionskip}{1pt}
\setlength{\belowcaptionskip}{1pt}
\varphi_j=\mathrm{softmax}(\mathop{{\Large \mathrm{avg}} }\limits_{m,n \in HW}(\bm{L}_P^j(m,n)))
\end{align}
where a larger value represents a greater contribution.

Finally, we perform a reweighted fusion of class prototypes $\bm{P}_f^j (\bm{P}_b^j)$ obtained by the CSRM module in each support branch with appearance factor $\varphi_j$, and use the fused prototypes to predict the final result, at which point Equation \ref{equation:25} becomes:
\begin{align}
\setlength{\abovecaptionskip}{1pt}
\setlength{\belowcaptionskip}{1pt}
\bm{y}^{Final} = \mathrm{softmax}({{\mathcal{F}}_{cosine}}(\bm{F_{q}},\sum_{j=1}^{\mathrm{\bm{K}}} \varphi_j \bm{P}_f^j,\sum_{j=1}^{\mathrm{\bm{K}}} \varphi_j \bm{P}_b^j)) 
\end{align}

\section{Experiments}\label{sec:Experiments}

First, we describe the setting of the experiment, including the experimental dataset, implementation details, and evaluation metrics in Section \ref{sec: Experimental Setting}. To demonstrate the effectiveness of our method, extensive comparative experiments are then performed with existing methods in Section \ref{sec: Comparison with State-of-the-arts}. And in Section \ref{sec: Ablation Studies}, we perform ablation experiments to analyze our DMNet. Then, we also provide statistical analyses in terms of different categories and scales in Section \ref{sec: Statistical Analysis}. Finally, some failure cases are described in Section \ref{sec: Failure case analysis}.

\subsection{Experimental Setting}\label{sec: Experimental Setting}
\subsubsection{Datasets}\label{sec:Datasets} 

To better validate the generalizability of the model to unknown classes, we prefer to choose the datasets with more classes. Therefore, We evaluate the proposed approach on two publicly available remote sensing semantic segmentation datasets(i.e., iSAID \cite{waqas2019isaid} and LoveDA \cite{wang2021loveda}) and a widely used
FSS dataset namely PASCAL-$5^i$ \cite{shaban2017one}. 

The iSAID dataset is a large-scale dataset for evaluating instance segmentation and semantic segmentation algorithms, which contains 655,451 object instances from 2,806 high-resolution images. The iSAID contains 15 object categories such as `ship', `baseball diamond', and `plane'. Following \cite{yao2021scale}, we split the 15 classes into three folds, each fold contains 10 training classes and 5 testing classes, with the training and testing classes in each fold not intersecting to simulate the known and unknown classes. With this cross-division, the dataset can be fully utilized to validate generalization to unknown classes. The details of the class splits (Fold-i) are given in Table \ref{tab:1}. We randomly crop the original image into a 256$\times$256 size to further enlarge the number of annotated samples. Following the BAM \cite{lang2022learning}, We remove the training images containing the category targets of the testing set to avoid information leakage from the testing set. For each fold, we randomly select 1000 support-query image pairs for the performance evaluation during the testing phase.

The LoveDA dataset is an urban-rural domain adaptive ground cover dataset for evaluating semantic segmentation and unsupervised domain adaptation algorithms which contains 166,768 semantic objects from three distinct cities along with 5987 high-resolution images. There are 7 kinds of object categories in LoveDA including `road', `water', `barren', `forest', etc. We consider experiments on this dataset because the cross-domain scenario is more challenging and potentially applicable for FSS. In our experiments, we ignore the background category and use only the last 6 categories. Similar to iSAID, for LoveDA, we split the 6 classes into 3 folds. We randomly crop the original image into a 473$\times$473 size and remove some training images that contain testing classes.

The PASCAL-$5^i$ dataset consists of the PASCAL VOC 2012 \cite{everingham2010pascal} and additional SDS datasets \cite{hariharan2011semantic}. Following the BAM \cite{lang2022learning}, the 20 classes are divided into four folds and each containing 5 classes.
\begin{table*}[t]
\centering
\setlength{\belowcaptionskip}{2pt}
\caption{The class split settings of iSAID and LoveDA} \label{tab:1}
\renewcommand{\arraystretch}{1.2}
\resizebox{0.85\linewidth}{!}{
\begin{tabular}{cccc}
\toprule[1pt]
Datasets                & Fold-i & Training classes & Testing classes    \\ \hline
                        & Fold-0  &\begin{tabular}[c]{@{}c@{}}ground track field, bridge, large vehicle, small vehicle, helicopter \\ swimming pool, roundabout, soccer ball field, plane, harbor  \end{tabular} & \begin{tabular}[c]{@{}c@{}}ship, storage tank, baseball diamond, \\ tennis court, basketball court\end{tabular} \\ 
\cline{2-4}
{iSAID}                 & Fold-1  &\begin{tabular}[c]{@{}c@{}} ship, storage tank, baseball diamond, tennis court, basketball court \\ swimming pool, roundabout, soccer ball field, plane, harbor \end{tabular} & \begin{tabular}[c]{@{}c@{}}ground track field, bridge, large vehicle, \\ small vehicle, helicopter\end{tabular} \\ 
\cline{2-4}
                        & Fold-2  & \begin{tabular}[c]{@{}c@{}}ship, storage tank, baseball diamond, tennis court, basketball court \\ ground track field, bridge, large vehicle, small vehicle, helicopter \end{tabular} & \begin{tabular}[c]{@{}c@{}}swimming pool, roundabout,  \\soccer ball field, plane, harbor\end{tabular}          \\ \hline
\multirow{3}{*}{LoveDA} & Fold-0  & water, barren, forest, agriculture & building, road     \\ \cline{2-4}
                        & Fold-1  &building, road, forest, agriculture & water, barren      \\ \cline{2-4}
                        & Fold-2  &building, road,water, barren & forest, agriculture \\ \toprule[1pt]
\end{tabular}}
\end{table*}
\subsubsection{Implementation Details}\label{sec:Implementation Details}

The proposed model is implemented using the Pytorch \cite{paszke2019pytorch} framework on NVIDIA RTX 2080Ti GPUs. The model is trained end-to-end by the SGD optimizer, where the weight decay is set to 0.0001 and the momentum is set to 0.9. And we employ poly \cite{chen2017deeplab} strategy to adjust the learning rate during the training phase. For experiments on iSAID, the images are resized to 256$\times$256, and we set batch size, the total number of epochs, and the initial learning rate to 8, 200, and 0.005, respectively. For experiments on LoveDA, the images are resized to 473$\times$473, and we set batch size, the total number of epochs, and the initial learning rate to 8, 200, and 0.003, respectively. For experiments on PASCAL-$5^i$, the images are resized to 473$\times$473, and we set batch size, the number of epochs, and the initial learning rate to 8, 200, and 0.005, respectively.

The baseline of our model consists of the baseline of PFENet \cite{tian2020prior} and ASPP \cite{chen2017deeplab}, which aims to enhance the adaptability for multi-scale scenes of remote sensing images. For comprehensive comparisons of performance, three backbone networks, VGG-16, ResNet-50, and ResNet-101 \cite{simonyan2014very,he2016deep} are selected for experiments. Following PFENet \cite{tian2020prior}, all of these backbones are pre-trained on ImageNet \cite{deng2009imagenet}, and these network parameters are fixed during the training phase to maintain a degree of generalizability. For the hyperparameters $\lambda$, $\mu_1$, and $\mu_2$ of the proposed model, we set them to 0.8, 0.7, and 0.6. A detailed discussion of the hyper-parameter settings is shown in the subsequent ablation study. 

Considering the paucity of FSS models for remote sensing scenes, we have selected several models that have worked well in natural scenes in recent years for comparison. To maintain a fair experimental environment, we use a consistent data enhancement strategy, number of training epochs, number of random seeds, and batch size.
\subsubsection{Evaluation Metrics}\label{sec:Evaluation Metrics}

As a subclass of semantic segmentation (single-class segmentation), we choose the generic and representative class mean intersection over union (mIoU) as an evaluation metric. For each class, the IoU can be calculated by $\mathrm{IoU}=(\frac{\mathrm{T_{P}}}{\mathrm{T_{P}}+\mathrm{F_{P}}+\mathrm{F_{N}}})$, where $\mathrm{T_{P}}$, $\mathrm{F_{P}}$, and $\mathrm{F_{N}}$ denote the number of positive samples with true predictions, positive samples with false predictions, and negative samples with false predictions, respectively. The formulation of mIoU follows $\mathrm{mIoU}=\frac{1}{n}\sum_{i=1}^{n}\mathrm{IoU_i}$, where n is the number of classes in each fold (i.e., n = 5 for iSAID, n = 2 for LoveDA and n=5 for PASCAL-5$^i$). And In addition, following \cite{rakelly2018conditional,dong2018few}, we also adopt the foreground-background IoU (FB-IoU) as an evaluation metric, which ignores image classes and calculates the mean of foreground IoU and background IoU for all test images. Thus, the formulation of FB-IoU follows $\mathrm{FB}$-$\mathrm{IoU} =\frac{1}{2}\sum_{i=1}^{2}\mathrm{IoU_i}$, where n =2. We take the average of results on all folds as the final mIoU / FB-IoU. In addition, for a comprehensive comparison, Accuracy (Acc) is used as a measure of the number of correctly categorized pixels in the result. We averaged the accuracy for each category to obtain the final result (mAcc). Due to its objectivity and comprehensiveness, we take the mIoU as the leading evaluation indicator.

\begin{figure*}[htb]
	\setlength{\abovecaptionskip}{2pt}
	\centering
	\includegraphics[width=0.8\linewidth]{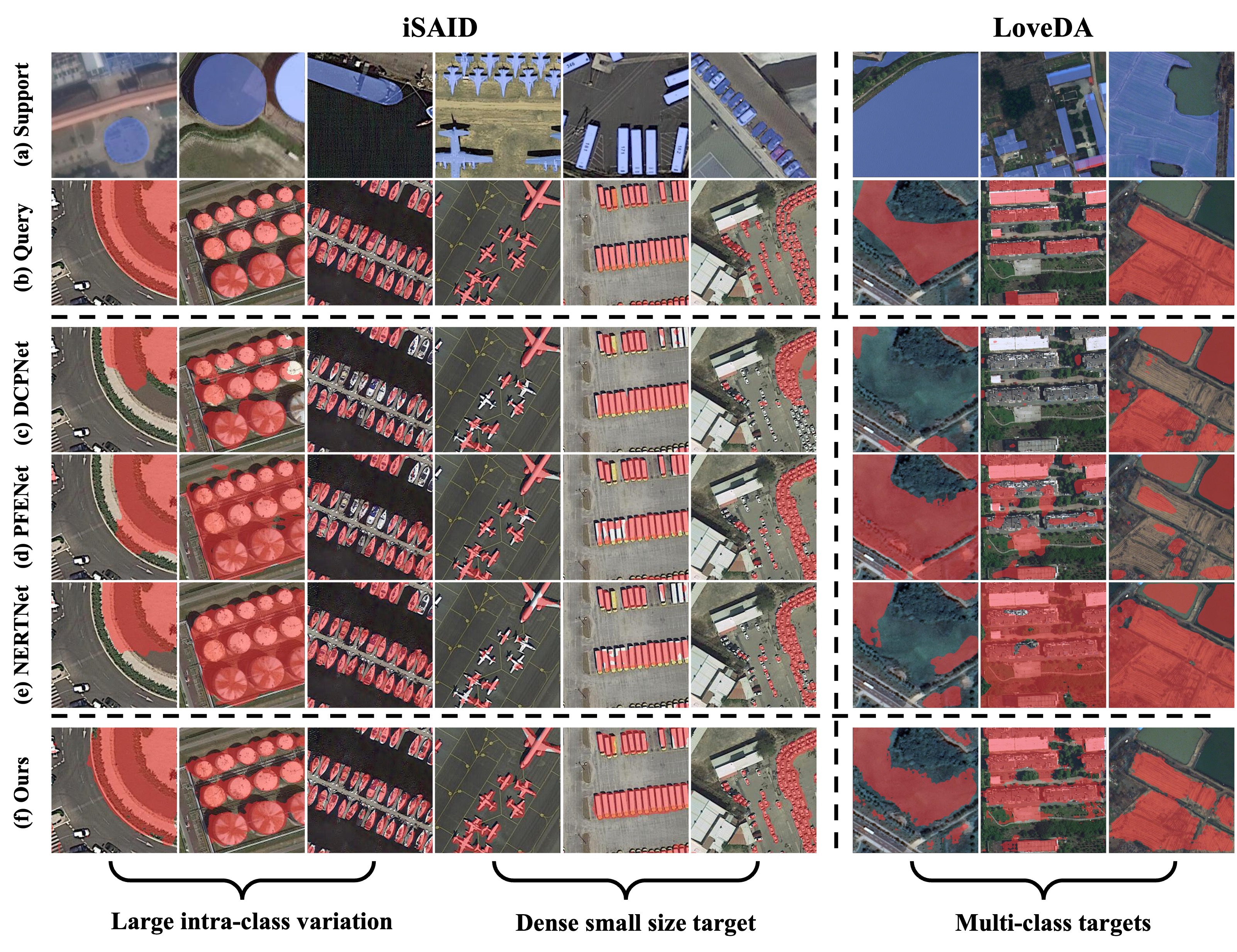}
	\caption{Qualitative results of our DMNet and some comparative models under the 1-shot setting. The left side column shows the results for iSAID, and the right side column shows the results for LoveDA. Each row from top to bottom : (a) support images with the ground-truth(GT) (\textcolor{blue}{blue}), (b) query images with the ground-truth(GT) (\textcolor{red}{red}), (c) results of DCPNet (\textcolor{red}{red}), (d) results of PFENet (\textcolor{red}{red}), (e) results of NTRENet (\textcolor{red}{red}), (f) results of our model (\textcolor{red}{red}).}
	\label{fig:6}
\end{figure*}

\begin{table*}[ht]
\setlength{\abovecaptionskip}{5pt}
\setlength{\belowcaptionskip}{10pt}
\caption{ Performance comparison on iSAID in terms of class mIoU and FB-IoU($\%$). The `Mean' denotes the averaged class mIoU scores for all three folds and the `FB-IoU' denotes the averaged class FB-IoU scores for all three folds. `Params' denotes the number of learnable parameters. \textbf{Bold} and \underline{underline} denote the best and second-best results, respectively.} \label{tab:2}
\renewcommand\arraystretch{1.2}
\centering
\resizebox{0.85\linewidth}{!}{
\begin{tabular}{ccccc|cc|ccc|ccc} 
\toprule[1pt]
\multirow{2}{*}{Backbone} & \multirow{2}{*}{Methond} &                            &                            & \multicolumn{1}{c}{1shot} &                                     & \multicolumn{1}{c}{} &                                     &                                     & \multicolumn{1}{c}{5shot} &                                     &                & \multirow{2}{*}{Params}                              \\ 
\cline{3-12}
                          &                          & Fold-0                     & Fold-1                     & Fold-2                    & Mean                                & FB-IoU               & Fold-0                              & Fold-1                              & Fold-2                    & Mean                                & FB-IoU         &                                     \\ 
\hline
  \multirow{7}{*}{VGG-16}                         & SADMNet \cite{yao2021scale}                 & 29.24                      & 20.80                      & 34.73                     & 28.26                               & -                    & 36.33                               & 27.98                               & 42.39                     & 35.57                               & -              & -                                   \\
                            & Jiang et al \cite{jiang2022few}                   & 33.40                      & 35.13                      & 43.28                     & 37.27                               & 57.72                & 37.33                               & 36.08                               & 44.52                     & 39.31                               & 59.01          & 14.7M                              \\
                          & SCLNet \cite{zhang2021self}                   & 46.88                      & 31.74                      & 44.64                     & 41.09                               & 60.88                & 47.75                               & 31.98                               & 43.45                     & 41.06                               & 60.69          & 11.5M                              \\
                          & PFENet \cite{tian2020prior}                  & 46.99                      & 35.33                      & 49.52                     & 43.95                               & 60.81                & \textbf{53.34}                      & \underline{38.55}                               & 50.61                     & \underline{47.50}                               & 62.94          & 10.4M                               \\
                    & ASGNet \cite{li2021adaptive}                   & 45.12                      & 32.25                      & 43.30                     & 40.22                               & 59.68                & 51.88                               & 33.93                               & \underline{50.96}                     & 45.59                               & \underline{63.07}          & 10.2M                               \\
                          & DCPNet \cite{lang2022beyond}                   & 47.44                      & 33.91                      & 48.08                     & 43.14                               & 60.43                & 48.16                               & 36.07                               & 49.25                     & 44.49                               & 62.03          & 25.3M                               \\
                          & NTRENet \cite{liu2022learning}                  & \underline{47.96}                      & \underline{36.21}                      & \textbf{51.22}            & \underline{45.13}                               & \underline{61.50}                & 49.05                               & 37.66                               & \textbf{52.33}            & 46.35                               & 62.24          & 20.5M                               \\ 

\rowcolor{gray!15} & DMNet                    & \textbf{50.64}             & \textbf{38.59}             & {46.32}                     & \textbf{45.18}                      & \textbf{63.10}       & \underline{52.09}                               & \textbf{40.54}                      & 50.39                     & \textbf{47.67}                      & \textbf{63.70} & \textbf{6.0M}                       \\ 
\hline
\multirow{8}{*}{ResNet-50}                         & SADMNet \cite{yao2021scale}                 & 34.29                      & 22.25                      & 35.62                     & 30.72                               & -                     & 39.88                               & 30.59                               & 45.70                     & 38.72                               &  -              & -                                   \\
                          & Jiang et al \cite{jiang2022few}                      & 35.11                      & 37.06                      & 42.24                     & 38.14                               & 58.28                & 38.19                               & 37.77                               & 43.94                     & 39.97                               & 59.13          & 25.6M                              \\
                          & SCL \cite{zhang2021self}                      & 49.08                      & 35.61                      & 48.15                     & 44.28                               & 61.66                & 50.69                               & 35.64                               & 48.72                     & 45.02                               & 62.63          & 11.9M                              \\
                          & PFENet \cite{tian2020prior}                   & \underline{51.34}                      & \underline{38.79}                      & 52.26                     & 47.46                               & \underline{63.34}                & \underline{54.71}                               & \textbf{41.51}                               & 54.45                     & \underline{50.22}                               & \underline{64.99}          & 10.8M                              \\
                  & ASGNet \cite{li2021adaptive}                   & 48.59                      & 36.82                      & 46.65                     & 44.02                               & 62.36                & 53.01                               & 37.44                               & 52.18                     & 47.54                               & 64.63          & 10.4M                               \\
                  
                          & CyCTR \cite{zhang2021few}                  & 51.15                      & 38.40                      & \textbf{53.79}            & \underline{47.78}                               & 62.86                & 51.91                               & 39.01                              & 54.83            & 48.58                               & 63.81          & 7.5M                               \\ 
                          & DCPNet \cite{lang2022beyond}                   & 48.43                      & 37.59                      & 52.09                     & 46.04                               & 62.76                & 50.34                               & 40.75                               & 51.33                     & 47.47                               & 64.10          & 35.0M                               \\
                          & NTRENet \cite{liu2022learning}                  & 49.52                      & 38.66                      & 51.87                     & 46.68                               & 62.84                & 50.60                               & 40.99                               & \underline{55.07}                     & 48.89                               & 63.74          & 20.8M                               \\ 
 
\rowcolor{gray!15}   & DMNet                    & \textbf{54.45} & \textbf{40.68} & \underline{53.60}            & {\textbf{49.58}} & \textbf{64.46}       & {\textbf{57.67}} & {\underline{41.06}} & \textbf{55.28}            & {\textbf{51.34}} & \textbf{65.81} & {\textbf{7.4M}}  \\ 
\hline
\multirow{7}{*}{ResNet-101}                           & SCL \cite{zhang2021self}                      & 47.59                      & 36.90                      & 45.21                     & 43.23                               & 61.71                & 48.94                               & 38.01                               & 46.21                     & 44.39                               & 60.93          & 11.9M                              \\
                          & PFENet \cite{tian2020prior}                  & 50.69                      & 38.37                      & \underline{52.85}                     & 47.30                               & 62.46                & \underline{54.40}                               & 41.55                               & 50.55                     & 48.83                               & 64.57          & 10.8M                               \\
                 & ASGNet \cite{li2021adaptive}                 & 47.55                      & 38.47                      & 49.28                     & 45.10                               & 62.03                & 53.54                               & 38.24                               & 53.20                     & 48.33                               & \underline{65.35}          & 10.4M                               \\
& CyCTR \cite{zhang2021few}                  & \underline{50.89}                      & \underline{38.89}                      & 52.22            & \underline{47.73}                               & 62.35                & 52.15                               & 40.28                              & \underline{55.32}            & \underline{49.25}                               & 64.45          & 7.5M                               \\ 
                          & DCPNet \cite{lang2022beyond}                 & 47.63                      & 38.80                      & 49.34                     & 45.26                               & 62.56                & 50.68                               & 40.02                               & 54.52                     & 48.41                               & 62.96          & 54.0M                               \\
                          & NTRENet \cite{liu2022learning}                 & 50.33                      & 38.73                      & 51.23                     & 46.76                               & \underline{63.25}                & 53.24                               &\textbf{41.87}                               & 51.53                     & 48.88                               & 64.16          & 20.8M                               \\ 
 \rowcolor{gray!15}
                          & DMNet                    & \textbf{54.01}             & \textbf{40.04}             & \textbf{53.57}            & \textbf{49.21}                      & \textbf{64.03}       & \textbf{55.70}                      & \underline{41.69}                      & \textbf{56.47}            & \textbf{51.29}                      & \textbf{65.88} & \textbf{7.4M}                       \\
                           \toprule[1pt]
\end{tabular}}
\end{table*}

\begin{table*}[ht]
\setlength{\abovecaptionskip}{5pt}
\setlength{\belowcaptionskip}{10pt}
\caption{ Performance comparison on LoveDA in terms of class mIoU and FB-IoU($\%$). The `Mean' denotes the averaged class mIoU scores for all three folds. And the `FB-IoU' denotes the averaged class FB-IoU scores for all three folds. `$\Delta$' represents the FB-IoU increment of segmentation performance under the 5-shot setting over the 1-shot setting. \textbf{Bold} and \underline{underline} denote the best and second-best results, respectively.} \label{tab:3}
\renewcommand\arraystretch{1.2}
\centering
\resizebox{0.85\linewidth}{!}{
\begin{tabular}{ccccc|cc|ccc|ccc} 
                           \toprule[1pt]
\multirow{2}{*}{Backbone}   & \multirow{2}{*}{Methond} &                &                & \multicolumn{1}{c}{1-shot} &                & \multicolumn{1}{c}{} &                &                & \multicolumn{1}{c}{5-shot} &                &             &\multirow{2}{*}{$\Delta$}
              \\ 
\cline{3-12}
          &         & Fold-0         & Fold-1         & Fold-2                     & Mean           & FB-IOU               & Fold-0         & Fold-1         & Fold-2                     & Mean           & FB-IOU      &    \\ 
\hline \multirow{6}{*}{VGG-16} 
          & SCL \cite{zhang2021self}     & 15.31          & 21.43          & 23.89                      & 20.21          & 29.56                & 14.84          & 22.39          & 20.85                      & 19.36          & 29.22      & -0.34     \\
          & PFENet \cite{tian2020prior}  & \underline{16.14}          & \underline{24.35}          & \underline{31.57}                      & \underline{24.02}          & 34.72                & 15.08          & \underline{26.82}          & 30.18                      & 24.03          & 38.40      &\underline{3.68}     \\
          & ASGNet \cite{li2021adaptive}  & 15.93          & 21.60          & 29.08                      & 22.20          & \underline{39.79}                & \underline{17.33}          & 26.24          & \textbf{35.75}             & \underline{26.44}          & \underline{38.41}        &-1.38   \\
          & DCPNet \cite{lang2022beyond}  & 15.22          & 22.58          & \textbf{31.83}             & 23.21          & 34.97                & 15.94          & 26.38          & 31.37                      & 24.56          & 36.24     &1.27      \\
          & NTRENet \cite{liu2022learning} & 15.07          & 23.17          & 28.68                      & 22.31          & 36.18                & 15.30          & 25.12          & 30.95                      & 23.79          & 35.06      &-1.12     \\ 
\rowcolor{gray!15} 
          & DMNet   & \textbf{19.71} & \textbf{26.23} & 30.43                      & \textbf{25.46} & \textbf{45.78}       & \textbf{25.02} & \textbf{35.51} & \underline{33.62}                      & \textbf{31.38} & \textbf{51.70}   & \textbf{5.92}\\ 
\hline \multirow{7}{*}{ResNet-50} 
          & SCL \cite{zhang2021self}     & 15.14          & 20.45          & 25.00                      & 20.20          & 24.60                & 14.25          & 21.09          & 23.65                      & 19.66          & 24.69       &0.09    \\
          & PFENet \cite{tian2020prior}  & \underline{17.13}          & 22.20          & \underline{26.49}                      & \underline{21.94}          & 33.48                & 15.83          & 25.73          & 24.74                      & 22.10          & 34.77      &1.29     \\
          & ASGNet \cite{li2021adaptive}  & 15.91          & 20.21          & 22.33                      & 19.48          & 36.39                & \underline{18.38}          & 26.29          & \textbf{36.34}                      & \underline{27.00}          & 39.59      &3.20     \\
          & CyCTR \cite{zhang2021few}  & 13.17          & \underline{23.43}          & 21.99                      & 19.53          & \underline{38.47}                & 13.81          & \underline{27.4}          & 26.15                      & 22.45          & \underline{42.71}       &\underline{4.24}    \\
          & DCPNet \cite{lang2022beyond}  & 16.67          & 23.10          & 24.44                      & 21.40          & 36.36                & 13.43          & 25.59          & 28.06                      & 22.36          & 33.89       &-2.47    \\
          & NTRENet \cite{liu2022learning} & 16.05          & 22.69          & 21.87                      & 20.20          & 32.67                & 15.79          & 24.94          & 23.42                      & 21.38          & 31.79     &-0.88      \\ 
\rowcolor{gray!15}
          & DMNet   & \textbf{19.29} & \textbf{25.52} & \textbf{31.53}             & \textbf{25.45} & \textbf{43.40}       & \textbf{24.62} & \textbf{33.80} & \underline{33.12}             & \textbf{30.51} & \textbf{50.93}   &\textbf{7.53}\\ 
\hline
\multirow{7}{*}{ResNet-101}  
          & SCL \cite{zhang2021self}     & 15.62          & 17.87          & 25.54                      & 19.68          & 31.03                & 14.96          & 20.26          & 24.02                      & 19.75          & 31.57      &0.54     \\
          & PFENet \cite{tian2020prior}  & 15.83          & \textbf{25.73} & 24.74                      & 22.10          & 35.65                & 15.62          & 24.37          & 26.64                      & 22.21          & 36.08      &0.43     \\
          & ASGNet \cite{li2021adaptive}  & 14.65          & 19.90          & 25.43                      & 19.99          & 30.89                & \underline{18.00}          & \underline{25.43}          & \underline{36.00}                      & \underline{26.48}          & \underline{39.26}       & \textbf{8.37}    \\
          & CyCTR \cite{zhang2021few}  & 13.16          & 20.63          & 20.55                      & 18.11          & \underline{38.94}                & 15.40          & 25.27          & 22.01                      & 20.89          & 36.03       &-2.91    \\
          & DCPNet \cite{lang2022beyond}  & \underline{16.52}          & 20.20          & \textbf{33.61}             & \underline{23.44}          & 36.57                & 16.97          & 25.08          & 25.10                      & 22.38          & 35.71    &-0.86       \\
          & NTRENet \cite{liu2022learning} & 15.51          & 19.65          & \underline{30.07}                      & 21.74          & 33.68                & 15.18          & 23.50          & 31.83                      & 23.50          & 32.16     &-1.52      \\ 
\rowcolor{gray!15}
          & DMNet   & \textbf{21.18} & \underline{24.57}          & 28.74                      & \textbf{24.83} & \textbf{47.47}       & \textbf{24.65} & \textbf{33.97} & \textbf{36.89}             & \textbf{31.84} & \textbf{53.38} &\underline{5.91} \\
                           \toprule[1pt]
\end{tabular}}
\end{table*}

\begin{table*}[htb]
\scriptsize
\setlength{\abovecaptionskip}{5pt}
\setlength{\belowcaptionskip}{10pt}
\caption{ Performance comparison on PASCAL-$5^i$ in terms of class mIoU ($\%$) with the backbone of ResNet-50. The `Mean' denotes the averaged class mIoU scores for all four folds. \textbf{Bold} and \underline{underline} denote the best and second-best results, respectively.} \label{tab:9}
\renewcommand\arraystretch{1.1}
\centering
\resizebox{0.75\linewidth}{!}{
\begin{tabular}{ccccc|c|cccc|c} 
\toprule
\multirow{2}{*}{Methond} &\multicolumn{5}{c}{1shot}  & \multicolumn{5}{c}{5shot}                  \\ 
\cline{2-11}
& Fold-0 & Fold-1 & Fold-2 & Fold-3 & Mean  & Fold-0 & Fold-1 & Fold-2 & Fold-3 & Mean   \\ 
\hline
Finetuning \cite{shaban2017one} &24.90 &38.80 &36.50 &30.10 & 32.60 & - &-&-&-&-\\
OSLSM \cite{shaban2017one} &33.60 &55.30 &40.90 &33.50 &40.80 &35.90 &58.10 &42.70 &39.10 &43.90 \\
Chen Y et al \cite{chen2022semi}        & 53.60  & 62.90  & 57.80  & 51.30  & 56.40 & 65.30  & 71.20  & \textbf{71.30}  & 63.20  & \underline{67.75}  \\

ASGNet \cite{li2021adaptive} & 58.84  & 67.86  & 56.79  & 53.66  & 59.29 & 63.66  & 70.55  & 64.17  & 57.38  & 63.94  \\
PFENet \cite{tian2020prior} & 61.70  & 69.50  & 55.40  & 56.30  & 60.73 & 63.10  & 70.70  & 55.80  & 57.90  & 61.88  \\
CyCTR  \cite{zhang2021few}& \textbf{65.70}  & \underline{71.00}  & 59.50  & \textbf{59.70}  & \underline{63.98} & \textbf{69.30}  & \underline{73.50}  & 63.80  & \underline{63.50}  & 67.53  \\
DCPNet \cite{lang2022beyond} & 63.81  & 70.54  & \underline{61.16}  & 55.69  & 62.80 & 67.19  & 73.15  & \underline{66.39}  & \textbf{64.48}  & \textbf{67.80}  \\
\hline
Baseline                 & 58.61  & 69.49  & 54.02  & 50.13  & 58.06 & 61.66  & 70.72  & 54.61  & 56.71  & 60.93  \\ \rowcolor{gray!15}
DMNet                    & \underline{64.55}  & \textbf{72.52}  & \textbf{61.50}  & \underline{57.82}  & \textbf{64.10} & \underline{67.39}  & \textbf{74.01}  & 64.93  & 62.08  & 67.10 \\
\bottomrule
\end{tabular}}
\end{table*}

\subsection{Comparison with State-of-the-arts}\label{sec: Comparison with State-of-the-arts}

In this section, we compare existing FSS methods on two remote sensing datasets and a widely used FSS dataset namely PASCAL-$5^i$. Table \ref{tab:2}-\ref{tab:12} detail the performance. We then qualitatively analyze the existing methods and our DMNet, with the comparative results in Fig.\ref{fig:6} and \ref{fig:8}.

\subsubsection{Quantitative Analysis}\label{sec:Quantitative Analysis}
\ 
\newline \indent \textbf{iSAID: } 
Table \ref{tab:2} provides the performance between our DMNet and several representative methods on iSAID under the 1-shot and 5-shot settings. It is noticeable that methods that perform good performance in natural scenarios fail to continue to lead in remote sensing scenarios. A large part of the reason is that there are huge differences between remote sensing scenes and natural scenes. Compared with natural scenes, remote sensing scenes have the characteristics of large intra-class variance, and there are many complex scenes with dense small-size targets and multiple similar targets.

Nevertheless, our method addresses these problems better, outperforming other FSS methods by a significant margin, and setting new state-of-the-arts under all settings. Specifically, with the backbone of Resnet-50, our method achieves the averaged mIoU scores of 49.58{\%} and 51.35{\%} under the 1-shot setting and 5-shot setting, respectively, surpassing the state-of-the-art results by 1.8{\%} and 1.12{\%}. With the backbone of ResNet-101, we reach 1.48{\%} and 2.04{\%} mIoU improvements over CyCTR (2$^{nd}$ best in 1-shot and 5-shot), respectively, indicating the effectiveness and superiority of the DMNet, with similar advantageous results in the backbone of VGG-16.

% More specifically, our model almost achieves the best result at every Fold, especially in Fold-0, with far better accuracy than other models. For Fold-1, the reason why the accuracy is slightly lower is that there are similar categories in Fold-1 (e.g., test categories `small vehicle' and `large vehicle', training category `ship') that cause the model to have more false activation. However, our model performs almost the best performance. 

Meanwhile, in terms of FB-IoU, our model on three backbones achieves 1.6{\%}, 1.12{\%}, and 0.78{\%} increments over the previous best results under the 1-shot setting, respectively, with similar advantageous results under the 5-shot setting. Moreover, our model has the least number of learnable parameters due to the parameter-free module design, with only 7.4M using the backbone of ResNet-50 (one-fifth of that of DCPNet). In addition, as shown in Table \ref{tab:12}, our model with the ResNet-50 achieves 77.47{\%} mAcc under the 1-shot setting, surpassing the state-of-the-art results by 1.09{\%}, which fully demonstrates the effectiveness.

\textbf{LoveDA: } Table \ref{tab:3} presents the 1-shot and 5-shot results on LoveDA. It can be found that the mIoU metrics are generally low, which indicates that there are more false activation and incomplete predictions. Part of the reason is that this challenging multi-domain dataset expects better generalization of the model, which conflicts with the general over-fitting of FSS towards known classes, especially in remote sensing scenarios with multi-class coexistence and large intra-class variance.

Despite that, our model still performs excellent performance. Our model with the backbone of ResNet-50 beats the other best results by a considerable margin of 3.51{\%} and 3.51{\%} mIoU under the 1-shot and 5-shot setting, respectively, achieving similar benefits in other backbone networks. More surprisingly, for the FB-IoU metric, our model on three backbones achieves 13.29{\%}, 8.22{\%}, and 14.12{\%} increments over the previous best results under the 5-shot setting, respectively. This excellent performance illustrates that the proposed model has sufficient generalization capability to accommodate cross-domain scenarios. The incremental results `$\Delta$' from 1-shot to 5-shot show that with more samples, our model better captures the potential category representation information.

\textbf{PASCAL-$5^i$: } Table \ref{tab:9} presents the 1-shot and 5-shot results on PASCAL-$5^i$. It can be found that meta-learning-based methods have strong advantages compared to fine-tuning. Even though our proposed method is designed for remote sensing scenarios, it shows competitiveness on visual datasets as well. Specifically, our model with the backbone of ResNet-50 achieves 64.10{\%} mIoU and beats the other best results under the 1-shot setting. And under the 5-shot setting, our proposed DMNet is equally competitive. Moreover, we simply perform ablation experiments and the results show that our method achieves 6.04{\%} and 6.17{\%} mIoU increments under the 1-shot and 5-shot settings, respectively, compared to the baseline, which well illustrates the effectiveness and generalization of the proposed method.
\begin{table}[t]
\setlength{\abovecaptionskip}{5pt}
\setlength{\belowcaptionskip}{10pt}
\caption{Comparative study of model complexity}\label{tab:12}
\renewcommand\arraystretch{1.2}
\centering
\resizebox{0.9\linewidth}{!}{
\begin{tabular}{cccccc} 
\toprule
Method  & mIoU &mAcc & \begin{tabular}[c]{@{}c@{}}Params\\(MB)\end{tabular} & \begin{tabular}[c]{@{}c@{}}GPU Memory\\(GB)\end{tabular} & \begin{tabular}[c]{@{}c@{}}Inference Speed\\(FPS)\end{tabular}  \\ 
\hline
SCL \cite{zhang2021self}    & 44.28 &75.71 & 11.9                                                & 3.82                                                     & 10.59                                                           \\
PFENet\cite{tian2020prior}  & 47.46 &75.88 & 10.8                                                & 2.40                                                     & 16.08                                                           \\
ASGNet \cite{li2021adaptive} & 44.02 &74.00 & 10.4                                                 & 2.26                                                     & 14.06                                                           \\
CyCTR \cite{zhang2021few}  & 47.78 &76.11 & 7.5                                                  & 4.74                                                     & 10.36                                                           \\
DCPNet \cite{lang2022beyond} & 46.04 &74.63 & 35.0                                                   & 2.62                                                     & 15.48                                                           \\
NERTNet\cite{liu2022learning} & 46.68 &76.38 & 20.8                                                 & 2.82                                                     & 13.83                                                           \\ 
\hline \rowcolor{gray!15}
DMNet   & 49.58 &77.47 & 7.4                                                 & 2.95                                                     & 12.58                                                           \\
\bottomrule
\end{tabular}}
\end{table}

\textbf{Comparative study of model complexity: } In addition, we make a comparison of the complexity of the models in terms of the number of learnable parameters, GPU memory, and inference speed to the full extent. We used ResNet-50 as the backbone to conduct experiments under the 1-shot setting on iSAID, and to ensure a fair comparison, the Batch Size is uniformly set to 8 for all methods. As can be seen in Table \ref{tab:12}, the proposed method has the least number of learnable parameters (one-fifth of DCPNet). In addition, the proposed method is not of high computational complexity, and it occupies 2.95 GB memory during the training phase and achieves 12.58 FPS inference speed, which is 1.79 GB memory decreased and 2.22 FPS improved compared to CyCTR (2$^{nd}$ best on mIoU). Although there is a certain gap compared to PFENet which has the fastest inference speed, the proposed method has great superiority in segmentation performance and learnable parameters. In summary, the proposed method achieves the best segmentation accuracy with good model complexity.

\subsubsection{Qualitative Analysis}\label{sec:Qualitative Analysis}
\ 
\newline \indent \textbf{Comparison with other models: } 
To better analyze and comprehend the effect of models, we additionally visualize the corresponding segmentation results during the testing phase. We select some scenes with significant remote sensing characteristics (e.g., dense small-size targets, large intra-class variance, and multiple similar targets) for display, as shown in Fig.\ref{fig:6}. 

It can be found that advanced FSS models for natural scenes fail to predict effectively when faced with remote sensing scenes. In particular, for the `baseball field' (1$^{st}$ column) and `ship' (2$^{th}$ column), where the scale properties between the query image and the support image differ significantly, these models can only obtain incomplete prediction results. The reason is that the category information captured by these models from the support images is not representative and general enough to guide segmentation well. 
And there is a large number of missed detection for the dense small size `plane' (4$^{th}$ column) and `small vehicle' (6$^{th}$ column). Moreover, for the `building' scenes (8$^{th}$ column) containing other similar categories (e.g., `forest') and `barren' scenes (9$^{th}$ column) containing other similar categories (e.g., `agriculture'), these models have a large number of false activation pixels. This is caused by the poor differentiation of the target class information captured by the model from other classes. In other words, the phenomenon of model bias towards known classes leads to wrong class activation.

According to the last row of visualization results, our proposed model has good prediction results in different remote sensing scenarios with more accurate target mask boundaries and less false activation. We are able to segment nicely the `baseball field' and the `storage tanks' with full boundaries, all the `ships', `planes', and `vehicles'. This indicates that our model is insensitive to support samples and captures representative category information well from various support samples. At the same time, our model can predict `waters', `buildings', and `barren' while generating fewer false category pixel activation. This shows that our model can effectively tackle the problems faced by FSS in remote sensing scenarios and demonstrates strong generalization and adaptability. 
\begin{figure}[htb]
	\setlength{\abovecaptionskip}{1pt}
	\centering
	\includegraphics[width=0.75\linewidth]{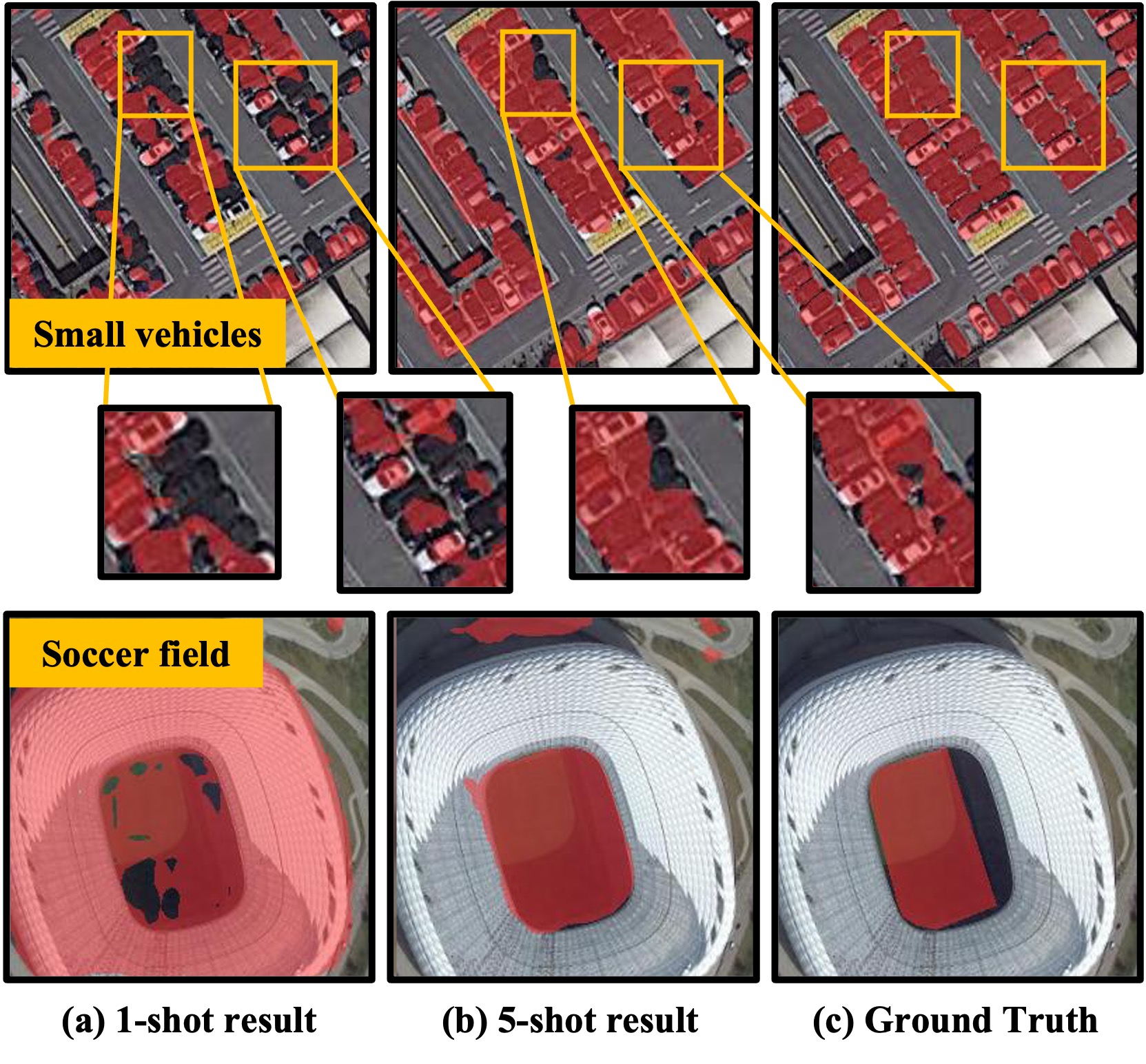}
	\caption{Comparison of 1-shot and 5-shot segmentation results. With more samples, our model produces more precise activation (more complete mask boundaries and fewer false activation).}
	\label{fig:8}
\end{figure}

\textbf{5-shot: } Furthermore, we also visualized the effect of the 5-shot setting on the segmentation results as shown in Fig.\ref{fig:8}. For the dense `small vehicles', the model under the 1-shot setting has some missed detection (the area enlarged in the $2^{nd}$ row), while the model can segment the target pixels better under the 5-shot setting. And for `soccer fields' where similar non-target pixels exist, the model can produce fewer false pixel activation under the 5-shot setting. It can be found that the proposed DMNet is able to capture more representative potential information of the target category given multiple support images. 

\begin{table}[t]
\setlength{\abovecaptionskip}{5pt}
\setlength{\belowcaptionskip}{10pt}
\caption{ Ablation studies of DMNet on iSAID under the 1-shot setting. `CPRM' denotes the Class-public Region Mining. `CSRM' denotes the Class-specific Region Mining. `KMS' denotes the Known-class Meta Suppressor. `Params' means the number of learnable parameters. `$\Delta$' represents the increment of variants of DMNet over the baseline (mIoU)} \label{tab:4}
\renewcommand\arraystretch{1.15}
\centering
\resizebox{0.9\linewidth}{!}{
\begin{tabular}{ccccccc} 
\toprule
CPRM & CSRM & KMS & mIoU           & FB-IoU         & Params      &{$\Delta$} \\ 
\midrule
           &            &           & 45.71          & 61.73          & \textbf{6.2M}  & -  \\
\ding{51} &            &           & 47.47          & 63.81          & 7.4M   & 1.76\\
           & \ding{51} &           & 47.37          & 63.30          & 6.2M   & 1.66\\
           &            & \ding{51}& 47.12          & 62.73          & 6.2M   & 1.41\\\midrule
\ding{51} & \ding{51} &           & 48.52          & 62.98          & 7.4M    & 2.81 \\
\ding{51} &            & \ding{51}& 47.60          & 63.15          & 7.4M    & 1.89\\
           & \ding{51} & \ding{51}& 48.59          & 64.42          & 6.2M    & 2.88\\
\midrule 
\ding{51} & \ding{51} & \ding{51}& \textbf{49.58} & \textbf{64.46} & 7.4M     & \textbf{3.87}\\
\bottomrule
\end{tabular}}
\end{table}
\subsection{Ablation Studies}\label{sec: Ablation Studies}
A series of ablation studies are performed in this section to explore the effect of each module on segmentation performance. Unless otherwise stated, the experiments are conducted on iSAID with the ResNet-50 backbone in this section.
 \subsubsection{Overview}\label{sec:Overview}

As shown in Table \ref{tab:4}, it can be found that the baseline with the CPRM module, CSRM module, and KMS module have different degrees of improvement compared to the baseline, showing that all three proposed components facilitate the segmentation. Additionally, the performance of the model improves when any two components are used, suggesting that our modules promote each other and work together to obtain better segmentation results rather than constrain each other. According to 5{$^{th}$} to 8{$^{th}$} rows, performance deteriorates in the absence of one of three modules, demonstrating the necessity of three modules for prediction. 

Our model ultimately achieves results of 49.58{\%} mIoU and 64.46{\%} FB-IoU by integrating three modules. Notably, the CSRM module does not use any learnable parameters, while the KMS module only needs to store the known class foreground and background prototypes (roughly 60KB) during the training phase. This allows our model to obtain a 3.87{\%} mIoU improvement by adding only 2.2M parameters. 

\begin{figure*}[ht]
	\setlength{\abovecaptionskip}{2pt}
	\centering
	\includegraphics[width=0.8\linewidth]{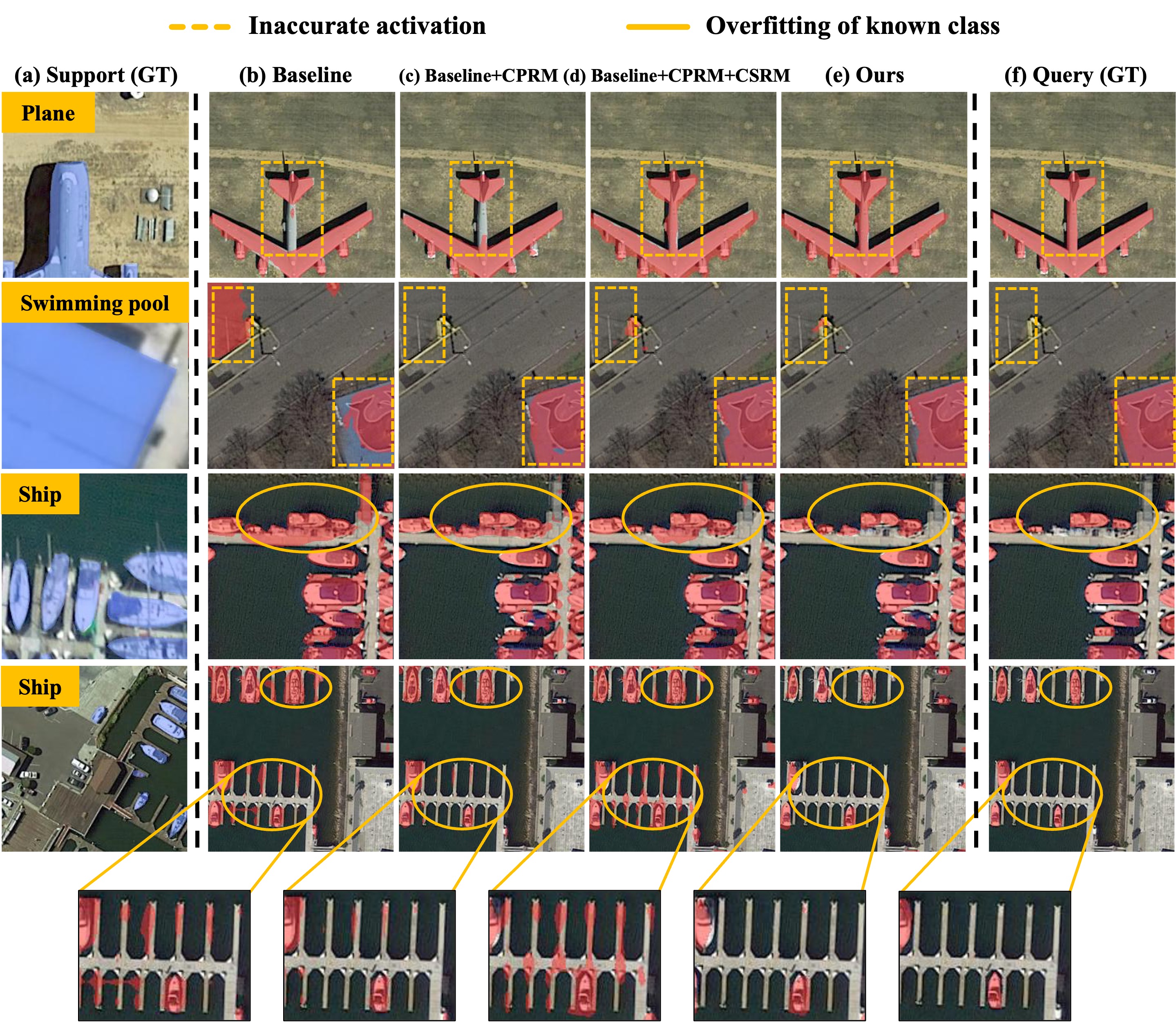}
	\caption{Qualitative visualization results of baseline and the variants of DMNet. From left to right: (a) support images with the ground-truth(GT) masks (\textcolor{blue}{blue}), (b) the results of the baseline (\textcolor{red}{red}), (c) the results of utilizing CPRM (\textcolor{red}{red}), (d) the results of utilizing CPRM+CSRM (\textcolor{red}{red}), (e) the results of utilizing CPRM+CSRM+KMS (i.e., DMNet) (\textcolor{red}{red}), (f) query images with the ground-truth(GT) masks (\textcolor{red}{red}). The yellow rectangular box indicates the typical region in the segmentation result.}
	\label{fig:7}
\end{figure*}

We visualize the results for the variants of DMNet and baseline under the 1-shot setting, as shown in Fig.\ref{fig:7}. It's remarkable that the baseline approach incorrectly activates some regions that are easily confused or belong to known classes, producing incomplete masks. 
% More specifically, for `ship' in the 1{$^{st}$} and 4{$^{th}$} rows, baseline tends to predict the known class `harbor' as foreground (yellow box); for `plane' in the 2{$^{nd}$} row and `swimming pool' in the 3{$^{th}$}, baseline produces incomplete predictions. 

However, the proposed DMNet produces more accurate predicted masks and fewer false activation than the baseline, suggesting that our model can capture representative class information and alleviate model bias toward known classes. More details on the effects of each module are provided in Section \ref{sec: The effect of CPRM module}-\ref{sec: The effect of KMS module}. 

\begin{figure}[h]
	\setlength{\abovecaptionskip}{1pt}
	\centering
	\includegraphics[width=0.9\linewidth]{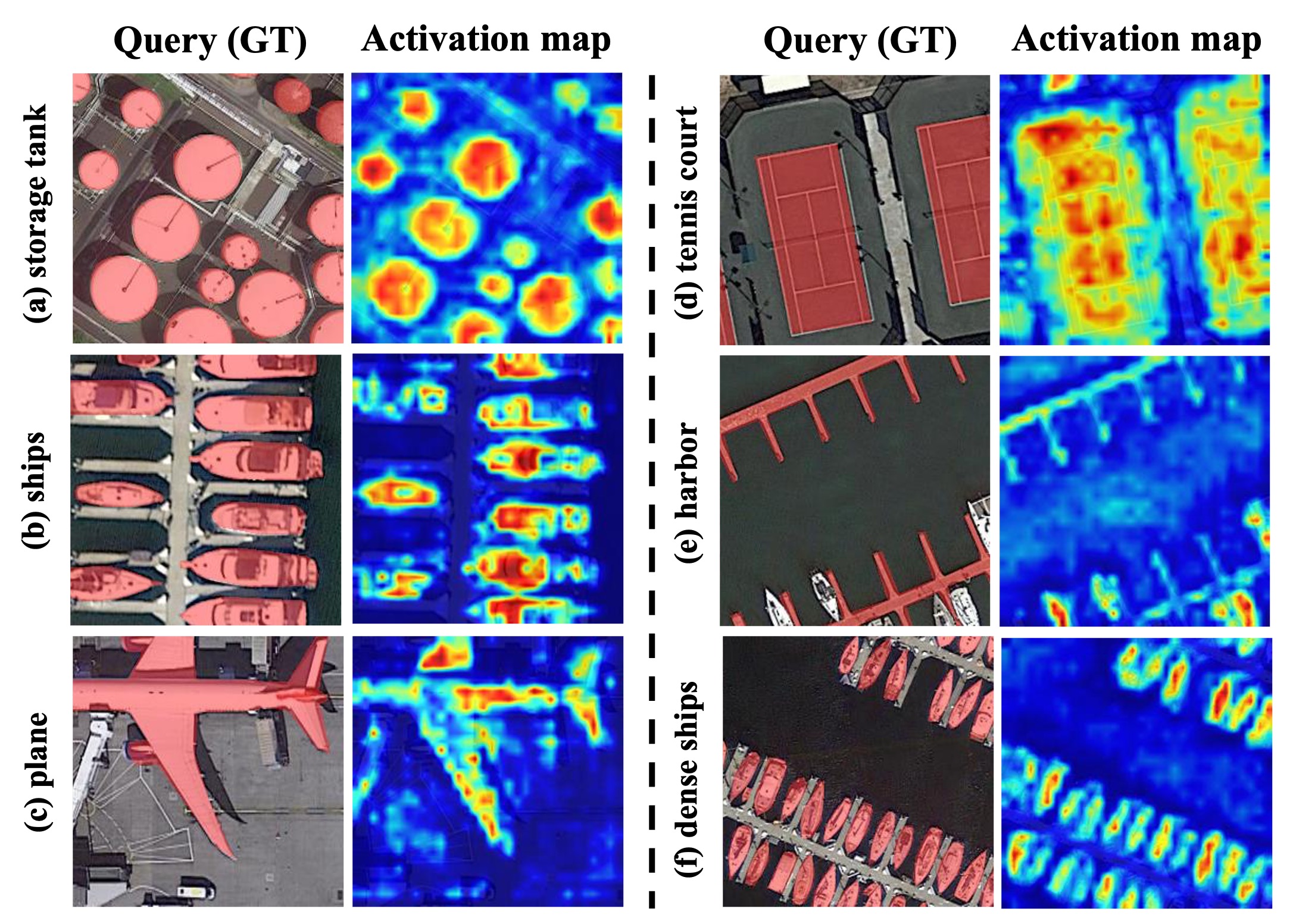}
	\caption{The visualization result (activation map) of the query feature through the CPRM module. Red in the activation map means important, and blue means unimportant. From left to right: (a) storage tank, (b) ships, (c) plane, (d) tennis court, (e) harbor, and (f) dense small-size ships.}
	\label{fig:9}
\end{figure}
\subsubsection{The effect of CPRM module}\label{sec: The effect of CPRM module}
This section focuses on the effectiveness of the CPRM module first, visualized as shown in Fig.\ref{fig:9}. Then we analyze the effect of different structures on performance, with the experimental results shown in Table \ref{tab:5}.

According to Section \ref{sec:Class-public Region Mining (CPRM)}, the CPRM module captures the bidirectional semantic association between the support-query image pair to reduce the distance between the image pair in the semantic space. 
As shown in the 3{$^{th}$} row in Fig.\ref{fig:7}, compared with the baseline, the CPRM module can segment a more complete `swimming pool' and produce almost no false pixel activation (yellow rectangular box area). This shows how well the module can handle the problem of large intra-class variance in remote sensing scenes. The 1.76{\%} increment of mIoU in Table \ref{tab:4} also verifies the effectiveness from the results. 

To further understand the functionality of the module, we visualize the query image passing through the CPRM module as shown in Fig.\ref{fig:9}. It can be found that after the module, the query image tends to activate the regions of the target category and suppress other regions, such as `storage tanks', `ships', and even for dense small-size `ships', all can be well activated.

We also perform ablation studies of the CPRM module with different structures as shown in Table \ref{tab:5}. It can be found that mining the class-public semantic only on position or on channel reduces mIoU by 0.29{\%} and 2.59{\%}, respectively, which indirectly explains the necessity of both PCRM and CCRM. It also shows that the category information is more concentrated on the positions than the channels, which is also consistent with the fact that people prefer to judge whether two objects are similar by their shapes. The results validate that the CPRM module can activate class-public regions in the query image similar to the target of the support image.
\begin{table}[t]
\setlength{\abovecaptionskip}{1pt}
\setlength{\belowcaptionskip}{1pt}
\caption{Ablation study of different structures in the CPRM module under the 1-shot setting. `Potision' denotes the Position-based Class-public Region Mining module (PCRM) and `Channel' denotes the Channel-based Class-public Region Mining module (CCRM).} \label{tab:5}
\renewcommand\arraystretch{1.0}
\centering
\resizebox{0.8\linewidth}{!}{
\begin{tabular}{ccccc} 
\toprule
Position & Channel & mIoU  & FB-IoU & Params  \\ 
\midrule                
\ding{51}        &         & 49.29 & 63.76  & \textbf{6.3M}    \\
         & \ding{51}       & 46.99 & 63.81  & 7.3M    \\
\ding{51}        & \ding{51}   & \textbf{49.58} & \textbf{64.46}  & 7.4M    \\
\bottomrule
\end{tabular}}
\end{table}

\subsubsection{The effect of CSRM module}\label{sec:The effect of CSRM module}
% The discussion first focus on the effectiveness of the CSRM module, as shown in Fig.\ref{fig:10}. The effect of different filter thresholds and Confusion-region Prototype Module (CPM) on performance is then explored, with the results presented in Table \ref{tab:6} and Table \ref{tab:14}.
The discussion first focuses on the effectiveness of the CSRM module, as shown in Fig.\ref{fig:10}. The effect of different filter thresholds on performance is then explored, with the results presented in Table \ref{tab:6}. Moreover, the generalisability of the CSRM module in other methods is discussed in Table \ref{tab:14}.

\begin{table}[t]
\setlength{\abovecaptionskip}{5pt}
\setlength{\belowcaptionskip}{0pt}
\caption{Ablation study of the CSRM module under the 1-shot setting. $\mu_1$ and $\mu_2$ denote confidence thresholds for foreground prototypes and background prototypes, respectively. `CPM' denotes the Confusion-region Prototype Module(CPM).} \label{tab:6}
\renewcommand\arraystretch{1.0}
\centering
\resizebox{0.65\linewidth}{!}{
\begin{tabular}{ccccc}
\toprule
$\mu_1$     & $\mu_2$    & CPM & mIoU          & FB-IoU          \\
\midrule  
0.6          & 0.6          & \ding{51}                & 48.70           & 64.26           \\
0.7          & 0.6          &                          & 48.60          & 63.96           \\
\textbf{0.7} & \textbf{0.6} & \ding{51}                & \textbf{49.58} & {64.46}  \\
0.8          & 0.6          & \ding{51}                & 49.41          & \textbf{64.51}           \\
0.8          & 0.7          & \ding{51}                & 49.03          & 64.32           \\
0.9          & 0.7          & \ding{51}                & 48.28          & 62.93           \\
0.9          & 0.8          & \ding{51}                & 48.62          & 63.75
\\ \bottomrule
\end{tabular}}
\end{table}

As mentioned in Section \ref{sec: Class-specific Region Mining (CSRM)}, the CSRM module aims to mine the category semantics of the query image itself to cross the intra-class variance gap. According to the 4{$^{th}$} and 6{$^{th}$} rows of Table \ref{tab:4}, it can be found that the module can improve the segmentation accuracy without using any parameters. To qualitatively analyze the role of the module, we visualize the results as shown in Fig.\ref{fig:10}. It can be found that more accurate target regions are obtained, such as the `planes' with extremely accurate activation areas, `tennis courts' and `storage tanks' with less false activation areas, which is a good indication of the importance and effectiveness of mining the semantics from the query image itself.
% As shown in column 5{$^{th}$}, the CSRM module can also utilize rich category semantics specific to the query image itself instead of just the support image semantics to produce more accurate predictions.

\begin{figure}[t]
	\setlength{\abovecaptionskip}{2pt}
	\centering
	\includegraphics[width=0.9\linewidth]{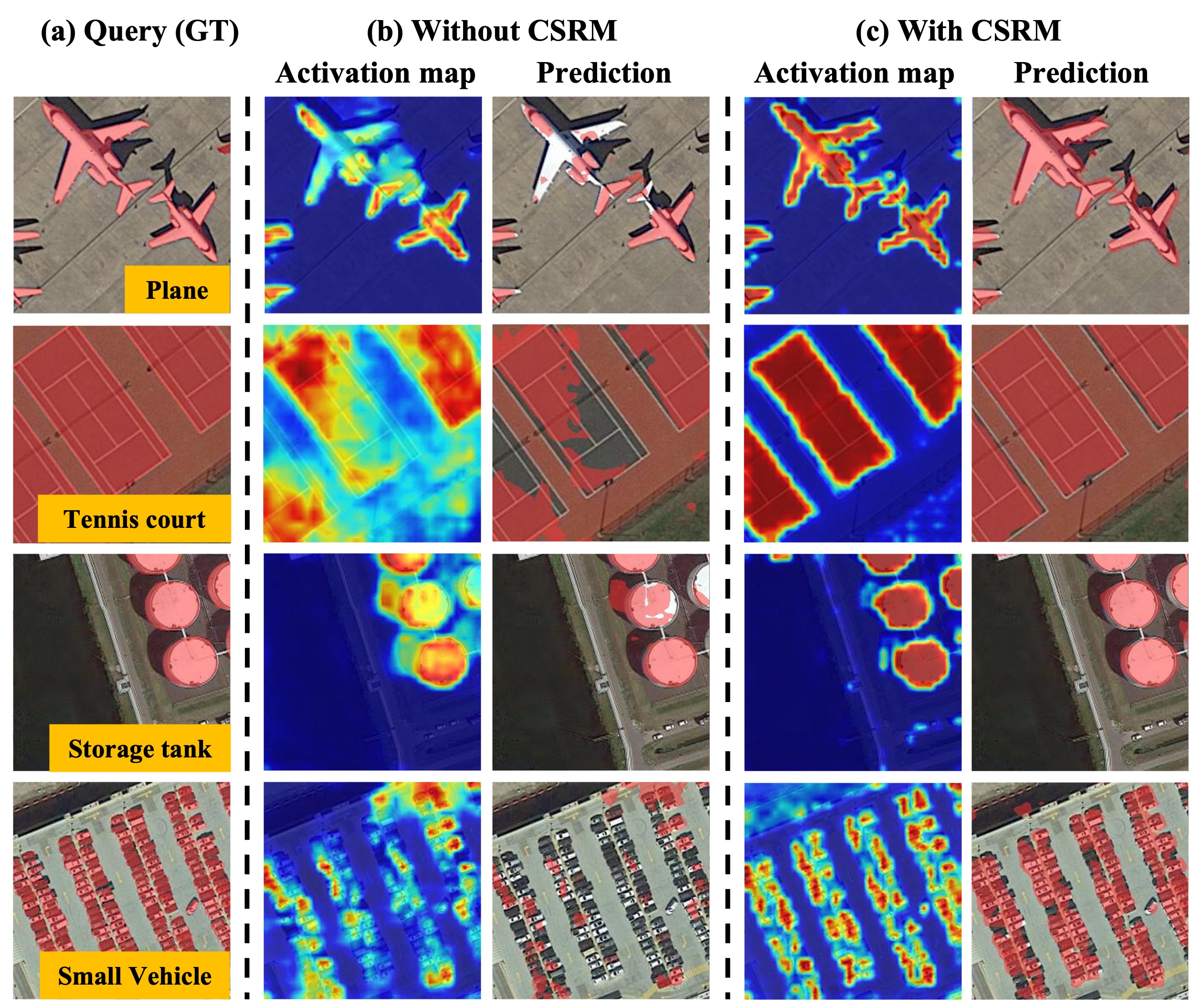}
	\caption{The visualization result of the CSRM module. (a) Query image (red represents the target area), (b) Activation map without CSRM module, (c) Prediction result without CSRM module, (d) Activation map with CSRM module, (e) Prediction result with CSRM module.}
	\label{fig:10}
\end{figure}

\begin{table}[t]
\setlength{\abovecaptionskip}{5pt}
\setlength{\belowcaptionskip}{0pt}
\caption{Generalisability study of the CSRM module in other methods under the 1-shot setting. \textcolor{blue}{Blue} color indicates the increments after connecting the CSRM module in series.} \label{tab:14}
\renewcommand\arraystretch{1.2}
\centering
\resizebox{0.75\linewidth}{!}{
\begin{tabular}{ccc} 
\toprule
Method       & mIoU        & FB-IoU       \\ 
\hline
SCL \cite{zhang2021self}           & 44.28       & 61.66        \\
 SCL+CSRM      & 46.25$_{\textcolor{blue}{(+1.97)}}$ & 63.11$_{\textcolor{blue}{(+1.45)}}$   \\ 
\hline
 PFENet  \cite{tian2020prior}       & 47.46       & 63.34        \\
 PFENet+CSRM   & 48.48$_{\textcolor{blue}{(+1.02)}}$ & 64.23$_{\textcolor{blue}{(+0.89)}}$  \\ 
\hline
 Baseline      & 45.71       & 61.73        \\
 Baseline+CSRM & 47.37$_{\textcolor{blue}{(+1.66)}}$ & 63.30$_{\textcolor{blue}{(+1.57)}}$   \\
\bottomrule
\end{tabular}}
\end{table}

We explore the ablation studies of the CSRM module as shown in Table \ref{tab:6}. $\mu_1$ and $\mu_2$ control the selection of foreground and background prototypes of the query image. A higher threshold value means a higher confidence level for the selected features. Although we expect to select foreground prototypes with features of high confidence, the CPM module can improve the representation of prototypes by mining class-specific semantics from low confidence features, so the choice of $\mu_1$ is not sensitive to the results. And since the background features in the query image are cluttered and complex, it is logical to tolerate more noise and choose a threshold with lower confidence. Based on the experimental results, we believe that threshold $\mu_1$ and $\mu_2$ of 0.7 and 0.6, respectively, are appropriate choices. 

In addition, due to its plug-and-play nature, the CSRM module can be well integrated with other methods. Table \ref{tab:14} demonstrates the generalisability of the CSRM module in other methods. It can be found that after connecting the CSRM module in series, SCLNet and PFENet obtain 1.97{\%} and 1.02{\%} mIoU improvement, respectively, which fully demonstrates the generality and effectiveness of the CSRM module.

\begin{figure}[t]
	\setlength{\abovecaptionskip}{1pt}
        \setlength{\belowcaptionskip}{1pt}
	\centering
	\includegraphics[width=0.9\linewidth]{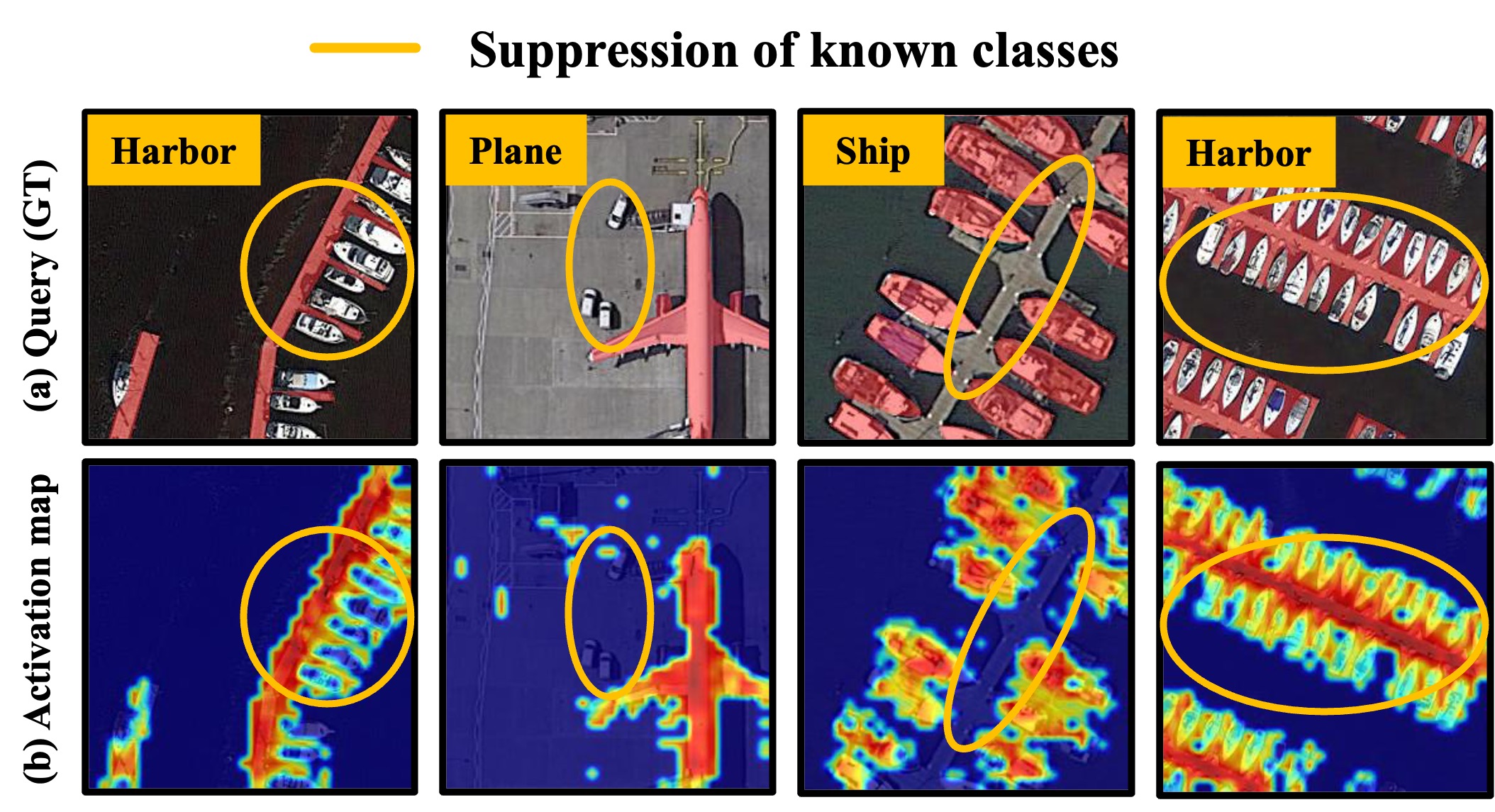}
	\caption{The visualization result of the KMS module. The KMS module enables better activation of the regions of unknown class while suppressing known class (e.g., `ship', `small vehicle', and `harbor').}
	\label{fig:12}
\end{figure}
\subsubsection{The effect of KMS module }\label{sec: The effect of KMS module}
\begin{figure}[t]
	\setlength{\abovecaptionskip}{1pt}
	\centering
	\includegraphics[width=0.9\linewidth]{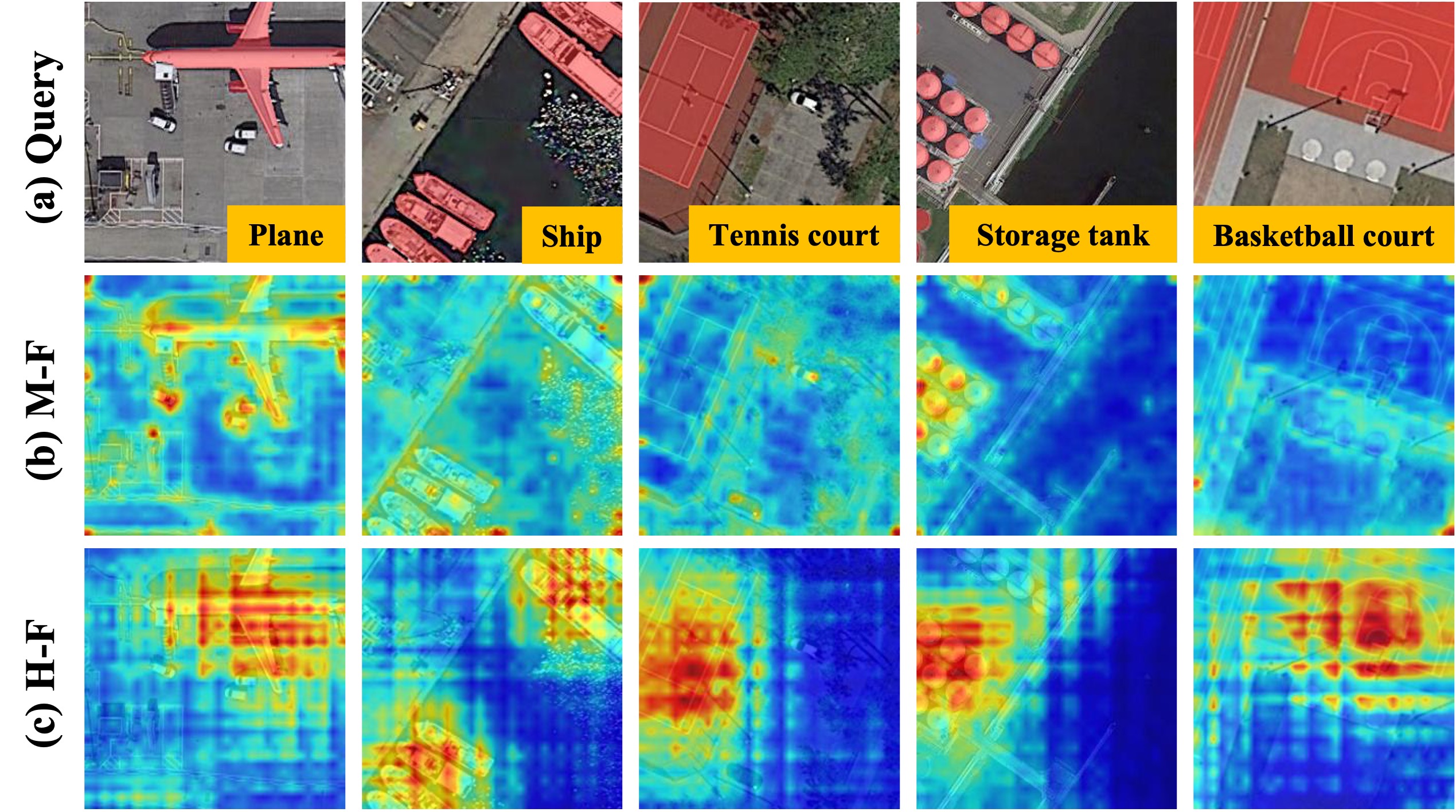}
	\caption{The visualization results for different levels of features (activation map). \textbf{Query}: query image, \textbf{M-F}: Middle-level feature, \textbf{H-F}: High-level feature. High-level features reveal more accurate regions of interest for target categories than mid-level features.}
	\label{fig:13}
\end{figure}

Firstly, we focus on the effectiveness of the KMS module, with the results demonstrated in Fig.\ref{fig:12}. Subsequently, we perform the ablation studies of different layers and sources of the category features captured during the training phase, whose results are presented in Fig.\ref{fig:13} and Table \ref{tab:7}. Finally, we explore the impact of hyper-parameter settings in the module.

To alleviate the generalization breakdown of unknown classes, we propose the Known-class Meta Suppressor (KMS) module by capturing the representative semantics of known classes and using them to suppress the activation of known classes in the query image. As shown in column 5{$^{th}$} in Fig.\ref{fig:7}, the KMS module almost suppresses the activation of the category `harbor', significantly alleviating the model bias toward known classes. We also visualize the activation maps of the KMS module as shown in Fig.\ref{fig:12}, which nicely demonstrates the role of the module in suppressing the activation of training known classes. As shown in column 5{$^{th}$} of Table \ref{tab:4}, the mIoU and FB-IoU of the model decreased by 1.06{\%} and 1.48{\%}, respectively, after removing the module, which also quantitatively proves the effectiveness from the results. 

Fig.\ref{fig:13} demonstrates that the high-level features are more class-specific and can locate more accurate target category regions than the mid-level features, hence the choice to mine high-level features instead of mid-level features. Table \ref{tab:7} also shows that better segmentation performance is obtained using higher-level features in terms of both mIoU and FB-IoU metrics.
\begin{table}[t]
\setlength{\abovecaptionskip}{5pt}
\setlength{\belowcaptionskip}{5pt}
\caption{Ablation study of the KMS module under the 1-shot setting. `FP' and `BP' represent the foreground prototypes and background prototypes of the unknown classes in the support images, respectively. `Block' represents the source of the captured feature information (mid-level or high-level).} \label{tab:7}
\renewcommand\arraystretch{1.0}
\centering
\resizebox{0.7\linewidth}{!}{
\begin{tabular}{ccccc} 
\toprule
FP & BP & Block  & mIoU           & FB-IoU          \\ 
\midrule
\ding{51}  &    & Middle & 48.66          & 63.27           \\
\ding{51}  & \ding{51}  & Middle & 49.06          & 63.82           \\
\ding{51}  &    & High   & 49.09          & 64.33           \\
\ding{51}  & \ding{51}  & \textbf{High}  & \textbf{49.58} & \textbf{64.46}  \\
\bottomrule
\end{tabular}}
\end{table}

Meanwhile, Table \ref{tab:7} also shows the effect of different sources for class prototypes on performance. It can be found that additionally using the background prototype of the support image in the Known-class Suppression Module to suppress the activation of non-target regions can lead to an increase in mIoU and FB-IoU by 0.49{\%} and 0.13{\%}, respectively. This is a good indication that there is some similarity in the background semantics in images of the same category, which can assist in activating the target regions of the novel class.
\begin{figure}[t]
	\setlength{\abovecaptionskip}{1pt}
	\centering
	\includegraphics[width=0.8\linewidth]{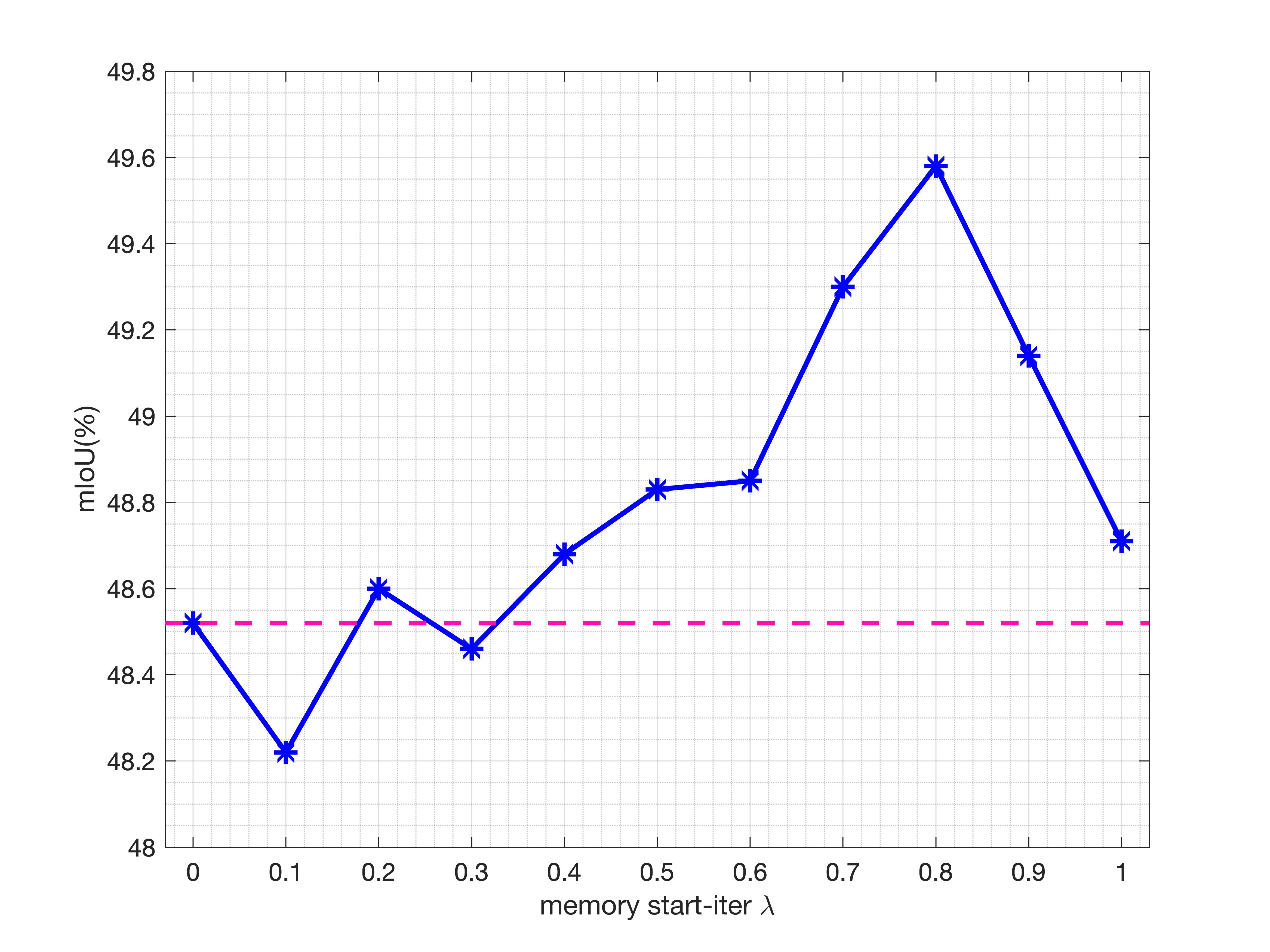}
	\caption{Study of the impact of the number of iterations of model start updates on performance in the KMS module. $\lambda$ represents that the model starts to update at the number of iterations of a certain epoch. For example, if the $\lambda$ is 0.2, and there are 1000 iterations in 1epoch, the model starts gradient update at 0.2*1000 iterations. Special note, here $\lambda$ is 0 means the KMS module is not used. The pink dashed line represents the segmentation accuracy without using the KMS module.}
	\label{fig:14}
\end{figure}

As mentioned in Section \ref{sec: KMS during the training phase}, the parameters of the model are updated only after a certain number of iterations to ensure the expressiveness of the known-class prototypes. We analyze the impact of the number of iterations the model starts to update on performance as shown in Fig.\ref{fig:14}. It can be found that the model undergoes more iterations ($\lambda$ from 0.1 to 0.8) to obtain better known-class prototypes and thus better segmentation performance, while the model undergoes few iterations to reduce the segmentation performance, which is basically consistent with our hypothesis. Also, when $\lambda$ continues to increase to 1 there is a certain decrease in the segmentation, which we believe that adopting a frozen backbone network for feature extraction will contain a certain amount of noise. After a certain number of iterations, some noise and errors may accumulate in the prototype and affect the incorrect prototype distribution. Therefore, based on the experimental results we set the hyper-parameter $\lambda$ to 0.8.

\begin{table}[ht]
\setlength{\abovecaptionskip}{5pt}
\setlength{\belowcaptionskip}{10pt}
\caption{Ablation studies of pre-trained visual models on iSAID in mIoU and FB-IoU.}\label{tab:10}
\renewcommand\arraystretch{1.2}
\centering
\resizebox{0.85\linewidth}{!}{
\begin{tabular}{ccccc} 
\toprule
\multirow{2}{*}{Backbone} & \multicolumn{2}{c}{1-shot} & \multicolumn{2}{c}{5-shot}  \\ 
\cline{2-5}
                          & mIoU  & FB-IoU             & mIoU  & FB-IoU              \\ 
\hline
VGG-16                    & 45.18 & 63.10              & 47.67 & 63.70               \\
ResNet-50                  & 49.58 & 64.46              & 51.34 & 65.81               \\
ResNet-101                 & 49.21 & 64.03              & 51.29 & 65.88               \\ 
\hline
ViT-B/16                  & 46.84 & 61.91              & 48.96 & 62.46               \\
Swin-B                    & 49.45 & 62.18              & 52.91 & 64.69               \\
\bottomrule
\end{tabular}}
\end{table}

\begin{table}[h]
\setlength{\abovecaptionskip}{5pt}
\setlength{\belowcaptionskip}{10pt}
\caption{Ablation studies of fine-tuning different layers of the backbone in the baseline on iSAID dataset}\label{tab:11}
\renewcommand\arraystretch{1.4}
\centering
\resizebox{0.95\linewidth}{!}{
\begin{tabular}{cccccc} 
\toprule
\multirow{2}{*}{Method}        & \multirow{2}{*}{Layer} & \multicolumn{2}{c}{1-shot} & \multicolumn{2}{c}{5-shot}  \\ 
\cline{3-6}
                               &                        & mIoU  & FB-IoU             & mIoU  & FB-IoU              \\ 
\hline
Fully finetune                 & 0,1,2,3,4              & 39.61$_{\textcolor{red}{(-6.10)}}$ & 59.74              & 43.62$_{\textcolor{red}{(-5.27)}}$ & 61.64               \\
\hline
\multirow{3}{*}{Part finetune} & 1,2,3,4                & 40.80$_{\textcolor{red}{(-4.91)}}$ & 59.22              & 45.87$_{\textcolor{red}{(-3.02)}}$ & 62.27               \\
                               & 2,3,4                  & 42.27$_{\textcolor{red}{(-3.44)}}$ & 60.09              & 45.98$_{\textcolor{red}{(-2.91)}}$ & 62.99               \\
                               & 3,4                    & 43.02$_{\textcolor{red}{(-2.69)}}$ & 59.48              & 47.76$_{\textcolor{red}{(-1.13)}}$ & 62.83               \\
\hline \rowcolor{gray!15}
Fix                            & -                      & 45.71 & 61.73              & 48.89 & 63.70               \\
\bottomrule
\end{tabular}}
\end{table}

\subsubsection{The effect of the backbone setting}\label{sec:The effect of the backbone setting}
In this section, we discuss the study of segmentation performance with different backbone settings.  

Firstly, we perform an ablation study on different pre-trained visual models. The experimental results are shown in Table \ref{tab:10}, it can be noticed that the performance with the backbone of ResNet-50 is instead better than the ResNet-101. Similar experimental findings were reported by PFENet \cite{tian2020prior} in their study. We argue that the adopted paradigm of freezing the backbone network may have inhibited the feature extraction effect of the deeper network. Moreover, considering that pre-trained large visual models have good generalization, we also explore their combination with FSS methods. Two transformers, Swin Transformer \cite{liu2021swin} and Vision Transformer \cite{dosovitskiy2020image} are chosen for validation. Following the existing training paradigm, we replace the previous CNN backbone with two pre-trained transformers and freeze them during the training phase to maintain a degree of generality. It can be seen that there is no expected additional gain after replacing the pre-trained large visual model compared to the previous CNN backbone, instead the performance decreases in most of the cases. Zhang et al.\cite{zhang2022feature} also failed to achieve the expected performance gain by using ViT \cite{dosovitskiy2020image} and Deit \cite{touvron2021training} as Backbone for CyCTR \cite{zhang2021few} and PFENet \cite{tian2020prior} in their transformer-based FSS study. Based on this phenomenon, we believe that the pre-trained large visual model may not match the current training paradigm of freezing backbone networks commonly adopted in FSS, and blindly selecting stronger backbone networks does not necessarily yield desirable results.

We then conduct a study of different fine-tuning strategies for the backbone. As shown in Table \ref{tab:11}, we adopt different fine-tuning strategies for ResNet-50 in the baseline, including full fine-tuning and part-fine-tuning. It is worth noting that i = 0, 1, 2, 3, 4 represents the fine-tuning of the i$^{th}$ layer parameter. It can be found that the more parameters of the backbone that are fine-tuned, the more severe the performance degradation compared to freezing the backbone, especially when fully fine-tuned, the mIoU scores decrease by 6.10{\%} and 5.27{\%} under the 1-shot and 5-shot settings, respectively. Similar conclusions were achieved by Sun et al.\cite{sun2022singular} in their experiments to fine-tune different backbone layers on Pascal-5$^i$. It shows that for the Few-Shot segmentation task, blindly fine-tuning the backbone can lead to bias toward known classes and affect the performance for segmenting unknown classes. Therefore, we choose to fix the backbone parameters to maintain a certain level of generalization.

\subsection{Statistical Analysis}\label{sec: Statistical Analysis}
To further explore the detailed impact mechanisms of our method, we statistically analyze the performance of our DMNet for different categories and different-scale objects in this section.
\subsubsection{The performance in different categories}\label{sec: The performance in different categories}

\begin{figure}[t]
	\setlength{\abovecaptionskip}{1pt}
	\centering
	\includegraphics[width=0.9\linewidth]{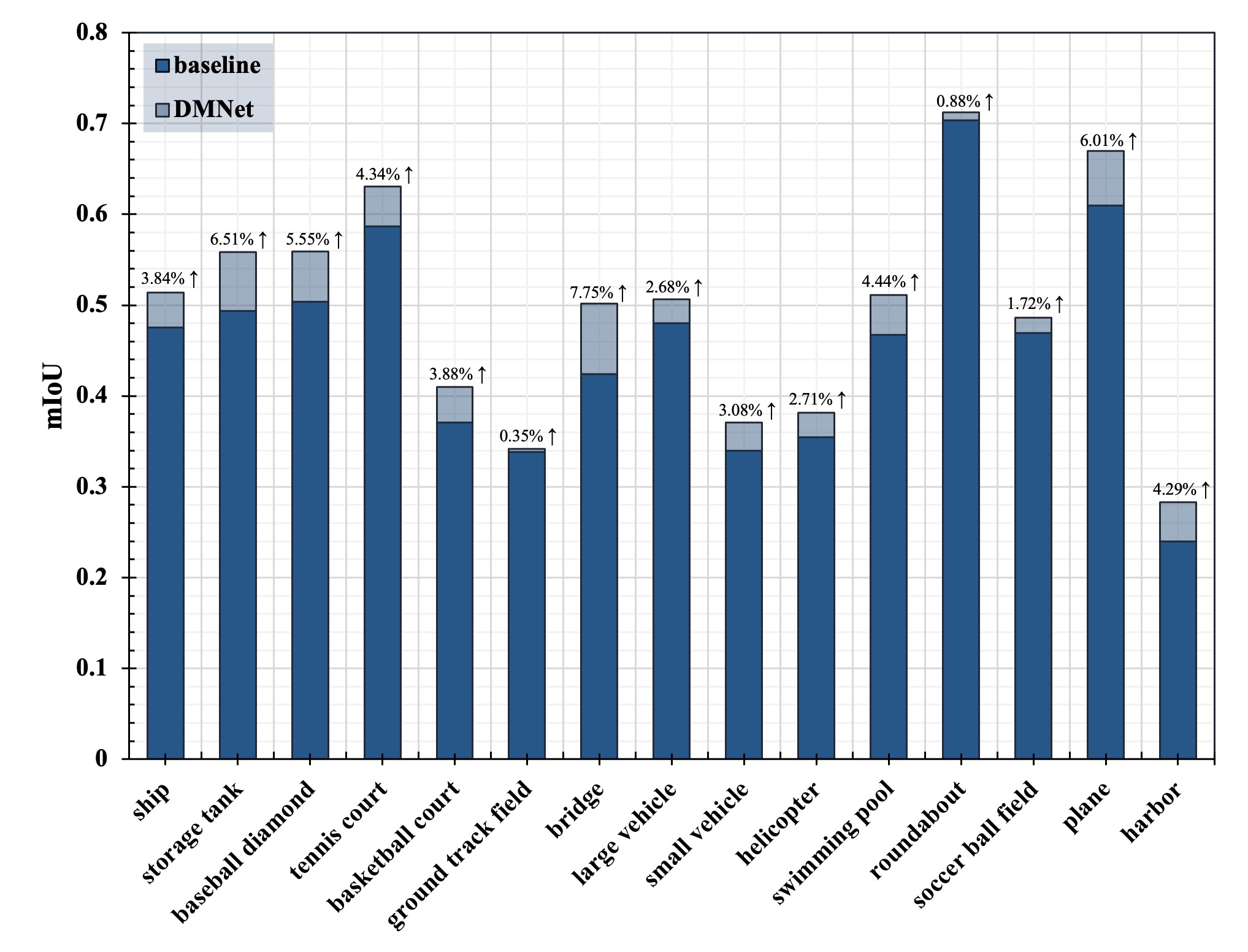}
	\caption{Performance comparison of DMNet and baseline in different categories.}
	\label{fig:15}
\end{figure}

Firstly, we compared the performance of the proposed DMNet with the baseline on different categories. Fig.\ref{fig:15} presents the results of experiments on the iSAID dataset. It can be found that the proposed DMNet achieves a good performance gain in each category. Among them, the `storage tank', `baseball diamond', `bridge', and `plane' categories gain more than 5{\%} mIoU increment. It is observed that the inter-class variance of these classes is large and varies greatly in the potential feature space, in which case the CPRM and CSRM modules in our DMNet can address the problem well from different perspectives. 

In addition, the baseline tends to confuse `ship' with `harbor', 'large vehicle' with 'small vehicle', and this over-fitting problem due to the model bias towards known classes can be well mitigated by the KMS module. 
% In particular, the `ground track field' category did not achieve significant gains compared to the other categories. The possible reason for this is that the number of target pixels in this category is often large or even covers the whole image, while our model is based on the prototype design, and it may be difficult to segment large areas of targets by calculating the similarity with only 256$\times$1$\times$1 feature vectors.
\subsubsection{The performance in different scales}\label{sec:The performance in different scales}

\begin{figure}[t]
	\setlength{\abovecaptionskip}{1pt}
	\centering
	\includegraphics[width=1.0\linewidth]{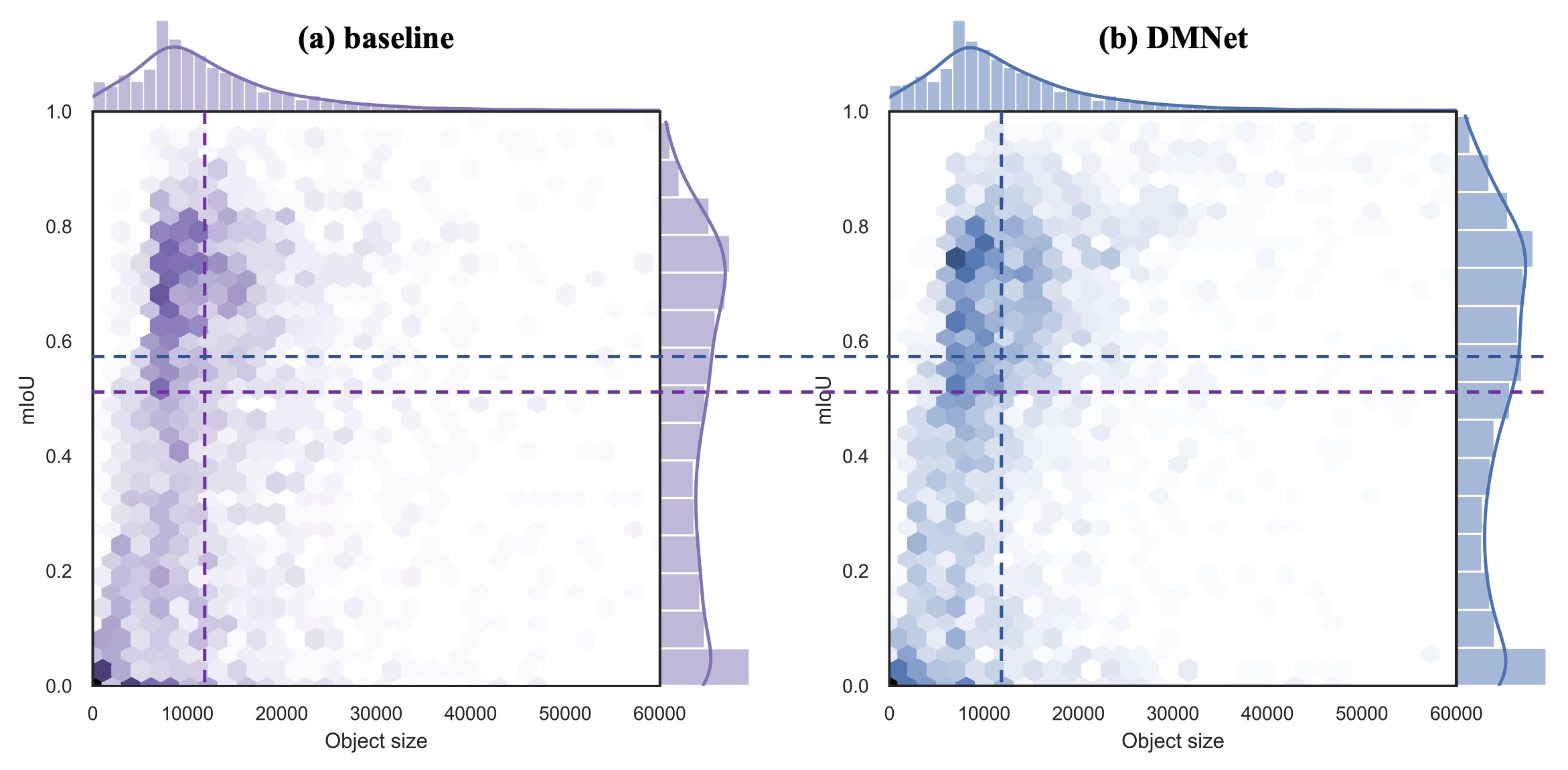}
	\caption{Performance comparison of DMNet and baseline in different scale objects. Here we consider the number of pixels in the target area as the object size (image size is 512$\times$512). It is clearly found that the proposed DMNet obtains a higher mIoU score compared to the baseline (the blue horizontal dashed line is above the purple horizontal dashed line).}
	\label{fig:16}
\end{figure}

To analyze the performance of the model for objects of different scales, we randomly sample 3000 query images of different classes in the iSAID dataset for 1-shot evaluation. Based on the sampling statistics of the object scales of the iSAID dataset as shown in Fig.\ref{fig:16}, we can see that most of the object scales are around 10,000 (the vertical dashed line represents the mean value of the scales). Subsequently, we analyze the segmentation performance and find that compared with the baseline; our DMNet obtains a higher mIoU (the horizontal dashed line represents the mIoU score), a larger number of high-quality segmentation results (top of the figure), and a smaller number of low-quality segmentation results (bottom left of the figure). These satisfactory results show that our model is strongly robust to object scale and can handle scale variations between the query and support images well, i.e., it can solve large intra-class variation problems.
\subsection{Failure case analysis}\label{sec: Failure case analysis}
\begin{figure}[t]
	\setlength{\abovecaptionskip}{1pt}
	\centering
	\includegraphics[width=0.85\linewidth]{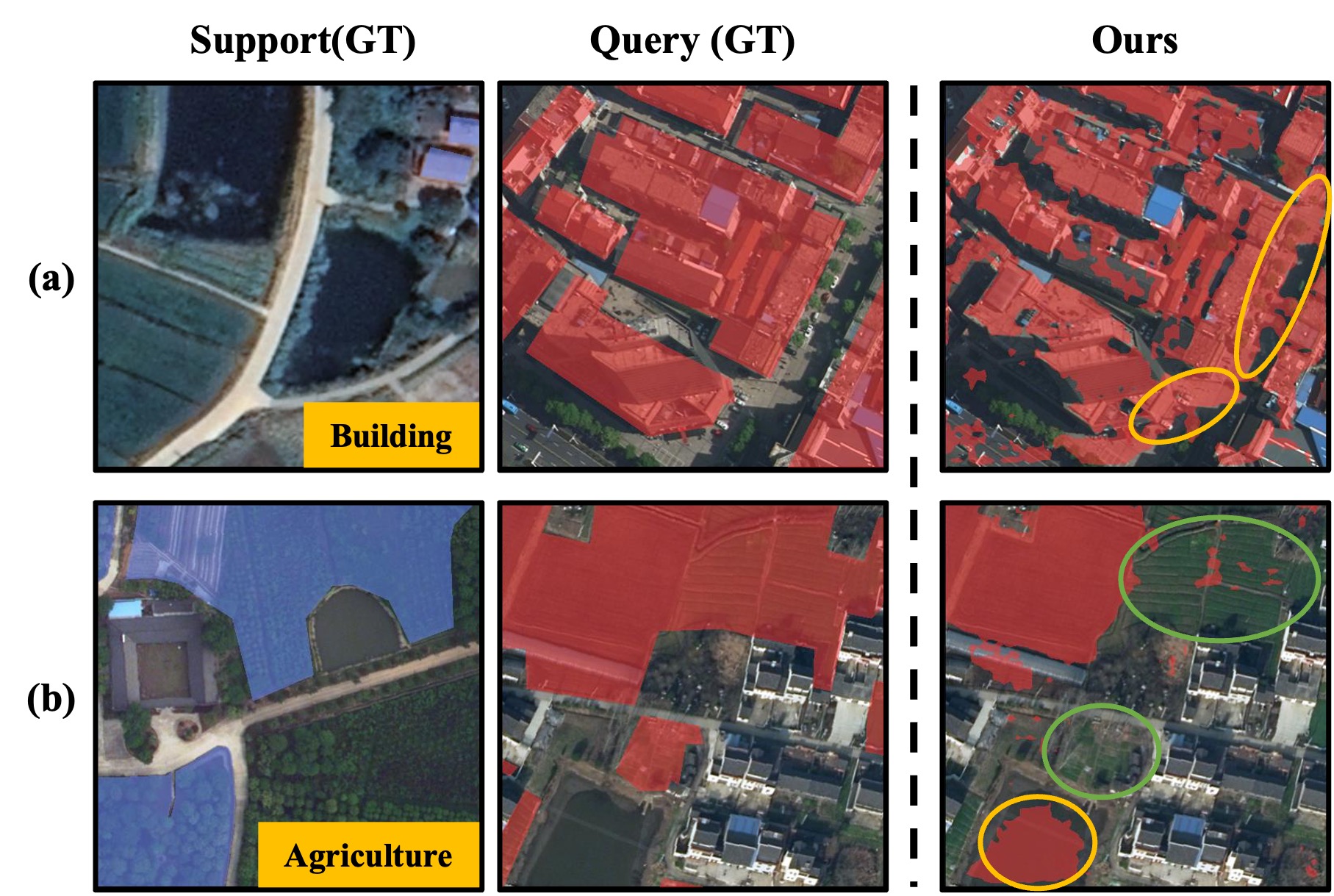}
	\caption{Some typical failure cases of the model. Specific false negatives and false positives are marked with green color and yellow circles.}
	\label{fig:17}
\end{figure}
Comparative experiments, ablation studies, and statistical analyses show that the proposed method achieves the best segmentation performance in two remote sensing datasets, iSAID and LoveDA. However, there are some issues that need to be addressed. As shown in Fig.\ref{fig:17}, when the target and background are extremely similar due to color, contrast, etc., there will be some false segmentation, e.g. the road in Fig.\ref{fig:17}(a) and the green pond in Fig.\ref{fig:17}(b) marked by the yellow circle. In addition, when there are multiple objects of the same class with large differences in appearance, texture, and color in the image, the proposed method is unable to focus on all the targets and has some misses, such as the targets marked by the green circle in Fig.\ref{fig:17}(b). We propose some thoughts that may help to solve these failures: (1) The proposed DMNet uses pixel-level cosine similarity computation to obtain predictions while ignoring local associations, and adding local semantic associations may be able to solve the failures. (2) Introducing a representative category prior to facilitating the model to better focus on the target region, e.g., semantically representative textual prompts. (3) Applying the foundation model with strong generalization performance to Few-shot tasks to better achieve Few-shot learning in remote sensing scenarios.
\section{Conclusion}
This paper proposes a DMNet network for few-shot segmentation in remote sensing scenes that no longer focuses solely on support images but pays more attention to the query image itself. To cross the intra-class variance gap, we propose the CPRM module and CSRM module to mine the public semantics of the target category and the target semantics specific to the query image itself, respectively, with both of them contributing to each other to jointly guide the segmentation. In addition, it is necessary to address the over-fitting of the model to the known classes. A unique meta-training paradigm is proposed to continuously learn the knowledge of known classes in a meta-memory, while the KMS module correspondingly utilizes the learned knowledge to suppress the activation of known class regions in unknown class samples. Extensive experiments on iSAID and LoveDA have validated the remarkable performance of DMNet, which outperforms previous approaches in achieving state-of-the-art performance with the minimum number of learnable parameters. In future work, we are committed to exploring the potential and feasibility of FSS in the multi-modal domain.

% \section{Declaration of competing interest}
% The authors declare that they have no known competing financial interests or personal relationships that could have appeared to influence the work reported in this paper.

% \section{Acknowledgments}
% This work for few-shot segmentation on Remote Sensing was supported by the National Natural Science Foundation of China (NSFC) under Grant 62171436.

% {\appendix[Proof of the Zonklar Equations]
% Use $\backslash${\tt{appendix}} if you have a single appendix:
% Do not use $\backslash${\tt{section}} anymore after $\backslash${\tt{appendix}}, only $\backslash${\tt{section*}}.
% If you have multiple appendixes use $\backslash${\tt{appendices}} then use $\backslash${\tt{section}} to start each appendix.
% You must declare a $\backslash${\tt{section}} before using any $\backslash${\tt{subsection}} or using $\backslash${\tt{label}} ($\backslash${\tt{appendices}} by itself
%  starts a section numbered zero.)}
%{\appendices
%\section*{Proof of the First Zonklar Equation}
%Appendix one text goes here.
% You can choose not to have a title for an appendix if you want by leaving the argument blank
%\section*{Proof of the Second Zonklar Equation}
%Appendix two text goes here.}

 % argument is your BibTeX string definitions and bibliography database(s)
%\bibliography{IEEEabrv,../bib/paper}
%
\bibliographystyle{IEEEtran}
\bibliography{IEEEabrv,refs}
\vfill

\end{document}